\documentclass[a4paper,fleqn]{cas-dc}

\usepackage[numbers]{natbib}
\usepackage{bbm}
\usepackage{multirow}
\usepackage{graphicx}
\usepackage{caption}

\usepackage{algorithm}
\usepackage{algpseudocode}

\algrenewcommand\algorithmicrequire{\textbf{Input:}}
\algrenewcommand\algorithmicensure{\textbf{Output:}}

\algrenewcommand\algorithmiccomment[1]{\hfill$\triangleright$~\footnotesize #1}

\usepackage{xcolor}
\definecolor{revblue}{RGB}{5,45,120}
\newcommand{\new}[1]{#1}

\def\tsc#1{\csdef{#1}{\textsc{\lowercase{#1}}\xspace}}
\tsc{WGM}
\tsc{QE}
\tsc{EP}
\tsc{PMS}
\tsc{BEC}
\tsc{DE}

\begin{document}
\let\WriteBookmarks\relax
\def\floatpagepagefraction{1}
\def\textpagefraction{.001}

\shorttitle{Intent Engine}

\shortauthors{Koushikur Islam et~al.}

\title [mode = title]{Intent Engine: Natural-Language Intent Translation for Intent-Driven Orchestration in the Compute Continuum}          




%
\author[1]{Koushikur Islam}[orcid=0009-0003-4619-7873]
\cormark[1]


\ead{k.shohag@westernsydney.edu.au}

\credit{Conceptualization, Investigation, Methodology, Data Curation, Software, Visualization, Writing - Original Draft}

\affiliation[1]{organization={Western Sydney University},
    city={Penrith},
    state={NSW},
    country={Australia}}

\author[1]{Rodrigo N. Calheiros}[orcid=0000-0001-7435-2445]
\ead{r.calheiros@westernsydney.edu.au}

\credit{Conceptualization, Methodology, Writing - Review \& Editing}


\cortext[cor1]{Corresponding author}



\begin{abstract}
\new{
Microservice placement in the compute continuum is driven by low-level Service-level Objectives (SLOs), but requiring users to specify metric-level constraints creates an adoption barrier and increases misconfiguration risk. Although large language models (LLMs) can interpret natural-language intents, direct generation of orchestration-consumable SLO artifacts remains unreliable due to unsupported constraints, incorrect grounded values, and schema violations. These errors can propagate to downstream placement logic and produce infeasible or incorrect placements. This paper presents \emph{Intent Engine}, a natural-language intent translation architecture that constructs validated SLO artifacts for compute-continuum service placement. \emph{Intent Engine} acts as an intent acquisition and SLO construction layer for existing intent-driven orchestration and placement frameworks; it does not perform placement or runtime QoS optimization. The architecture combines schema-constrained extraction, retrieval-grounded value construction from monitored infrastructure state, and validation against supported constraints before emitting the final SLO artifact. We evaluate \emph{Intent Engine} using a 716-record intent-to-SLO dataset derived from an edge--cloud testbed, including valid and invalid intents. Across GPT-4.1 mini, Claude Sonnet 4.5, and DeepSeek V4-Flash, \emph{Intent Engine} outperforms prompting baselines and a non-LLM rule-based parser. With GPT-4.1 mini, it achieves 0.941 total F1 Score and reduces aggregate hallucination by 85.1\%, while lowering downstream placement failure from 30.8\% to 2.1\%.

}
\end{abstract}


\begin{highlights}
\item Translates natural-language service placement intents into orchestration-consumable Service-level Objective (SLO) artifacts.
\item Provides retrieval-augmented and schema-bounded LLM-based intent-to-specification construction.
\item Reduces the need for specialized distributed-systems expertise in expressing service placement goals.
\item Supports adaptable integration with intent-driven placement algorithms and orchestration frameworks.
\end{highlights}



\maketitle
\section{Introduction}
\label{introduction}

Software applications have driven a paradigm shift toward \emph{microservices}, enabling improved scalability, flexibility, resilience, and faster delivery. By decomposing software into independently deployable components, microservices support dynamic scaling, rapid rollout, and reduced latency~\cite{sebrecht}. Modern microservice-based systems primarily rely on cloud infrastructures to achieve scalability, cost efficiency, and availability; however, centralized data centers introduce network latency due to limited geographic proximity to users~\cite{odun}. This limitation has led to the emergence of edge computing.

Edge computing~\cite{mansouri2021review} mitigates latency by processing data closer to its source in real time. Nevertheless, resource constraints at the edge restrict scalability, computational capacity, and availability compared to cloud environments. To balance these trade-offs, the \emph{compute continuum} integrates edge and cloud resources, enabling microservice deployment across heterogeneous infrastructures.

The compute continuum spans endpoints, edge, and cloud layers, distributing services to reduce latency while improving throughput and scalability~\cite{moreshi}. Within this environment, microservice placement becomes critical for meeting user requirements such as low latency, high availability, storage efficiency, and quality of service (QoS)~\cite{samani}. Applications including live streaming, AI inference, and digital advertising are particularly sensitive to these placement decisions.

Service-level Objectives (SLOs) are commonly used to guide placement across geographically distributed resources~\cite{samani,zafeiro,mota}. However, low-level SLO specification imposes a significant barrier for users lacking system expertise. Misconfigured placements can lead to QoS violations, service disruptions, and unexpected costs. Given the complexity and dynamic nature of continuum environments, accurate placement decisions are therefore essential for reliable operation.

Intent-driven orchestration (IDO) addresses this challenge by allowing users to express high-level goals, or intents, that describe desired outcomes rather than specific configurations~\cite{boutouchent2023amanos}. Inspired by Intent-Based Networking (IBN)~\cite{gharbaoui2023intent,clemm2022intent}, IDO autonomously translates these objectives into operational actions, abstracting low-level system details and reducing administrative burden. This abstraction lowers the expertise barrier and promotes scalable orchestration across heterogeneous \new{resources}.

Despite these benefits, existing orchestration frameworks~\cite{sebrecht, spillner, schahram, filinis2024intent, akbari2024icontinuum, metsch2023intent} typically accept intents as structured SLO specifications rather than unstructured natural language. Users must still provide precise metrics and system parameters, requiring domain knowledge and increasing the risk of misconfiguration, particularly under dynamic and heterogeneous conditions.

Recent advances in large pre-trained language models (PLMs) and large language models (LLMs) have demonstrated strong capabilities in natural language understanding and reasoning~\cite{min2023recent, he2024exploring}. These capabilities suggest that LLMs could translate human-readable intents into formal SLO specifications consumable by orchestration frameworks, thereby simplifying service placement for operators and DevOps practitioners.

However, employing LLMs to generate system-critical specifications introduces new risks. LLMs are prone to hallucinations, producing outputs that deviate from user intent or contextual correctness~\cite{bang2025hallulens}. In placement decisions, such errors can result in invalid configurations or system failures. Because continuum resources fluctuate dynamically, purely generative translations lack the guarantees required for reliable control.

In our prior work, \emph{MicroIntent}~\cite{islam2025microintent}, \new{we found that end-to-end LLM-based translation of natural-language placement intents can produce hallucinated or incorrect SLO constraints, even when supplied with contextual information}. \new{This is problematic in the compute continuum, where fluctuating resource states require reliable and grounded control-plane specifications rather than prompt-only generation}. These findings indicate that relying solely on end-to-end LLM inference can lead to misconfigurations, user intent violations, and unanticipated system impacts.

\new{Based on these insights, we argue that natural-language intent translation is a control-plane specification-construction problem that is different from typical generative language tasks. Current IDO frameworks require structured SLO inputs, whereas users often express placement intents as incomplete, ambiguous, or context-dependent details. A reliable acquisition layer must parse the intent, ground context-dependent requirements in monitored infrastructure state, check the result against the supported schema, and emit only validated SLO artifacts for downstream orchestration.}

\new{In this paper, we propose \emph{Intent Engine}, a natural-language intent translation architecture that converts service placement intents into platform-agnostic SLO specifications consumable by existing IDO and placement frameworks~\cite{sebrecht, spillner, schahram, filinis2024intent, akbari2024icontinuum, metsch2023intent}. \emph{Intent Engine} acts as an intent acquisition and SLO construction layer: it produces validated SLO artifacts to drive accurate intent-driven downstream orchestration. 

Since inaccurate translations can directly lead to infeasible or incorrect placements, our evaluation measures translation correctness, grounding accuracy, hallucination reduction, invalid-intent rejection, latency, scalability, and downstream placement impact. This paper makes the following contributions:}

\begin{itemize}
\new{
\item We introduce \emph{Intent Engine}, an intent acquisition system that converts natural-language service placement intents into validated, orchestration-consumable SLO artifacts.
}

\item We design a multi-stage SLO construction pipeline that separates semantic extraction, intermediate SLO representation, infrastructure-aware grounding, and schema validation.

\item We introduce infrastructure-aware grounding to resolve implicit constraints, such as highest or lowest resource values, from monitored compute-continuum state.
\new{
\item We evaluate \emph{Intent Engine} using constraint-level F1 Score, Exact Match, Jaccard similarity, hallucination rate, invalid-intent rejection, context retrieval ablation, latency, scalability, and downstream placement-failure impact.

\item We release a 716-record intent-to-SLO dataset from a real compute-continuum testbed, covering valid, ambiguous, conflicting, malformed, and unsupported intents.\footnote{\url{https://doi.org/10.5281/zenodo.20810799}}
}
\end{itemize}

\new{
Due to the language translation focus of \emph{Intent Engine}, tasks such as placement, deployment, or runtime QoS optimization are outside the scope of this paper.
}

The rest of the paper is organized as follows. Section~\ref{related_work} reviews existing work. Section~\ref{system_architecture} presents the \emph{Intent Engine} architecture. Section~\ref{implementation_and_experimental_setup} describes the implementation and compute-continuum testbed. \new{Section~\ref{evaluation} evaluates SLO construction accuracy, grounding effectiveness, hallucination reduction, failure handling, system overhead, and downstream placement impact. Section~\ref{limitations} discusses limitations} and Section~\ref{conclusion_and_future_work} concludes the paper with future work.

\section{Related Work}
\label{related_work}
\subsection{Intent-Driven Orchestration in the Compute Continuum}

Intent-Driven Orchestration (IDO) has emerged as a modern paradigm that enables decoupled application management through Service-level Objectives (SLOs), reducing administrative and operational overhead~\cite{metsch2023intent}. In IDO, intents express high-level policies or business goals, allowing users to specify \emph{what} the system should achieve while the orchestration logic autonomously determines \emph{how} to achieve it~\cite{santos2021towards, filinis2024intent}. Advances in Artificial Intelligence (AI) and Machine Learning (ML) further support this vision by enabling automated management of increasingly complex and dynamic compute continuum resources~\cite{asif2025leveraging}.

Despite these benefits, service placement across geographically distributed edge--cloud environments remains challenging due to fluctuating system states, resource trade-offs, and cost constraints that must still satisfy service intents~\cite{islam2025microintent}. Traditional Kubernetes-based orchestration~\cite{kubernetes_docs} lacks native intent-centric abstractions and often requires frequent manual intervention and specialized expertise, limiting its ability to adapt to dynamic continuum conditions.

Several studies extend Kubernetes and KubeEdge to automate deployment across the continuum~\cite{xiong2018extend, jansen2023continuum, al2024computing}. However, these solutions still rely on detailed low-level configuration and significant operator involvement. At enterprise scale, managing numerous interdependent services becomes increasingly complex, and placement decisions based on transient system states remain difficult even for experienced administrators.

\subsection{Intent Specification and Acquisition Approaches}

Traditional intent acquisition approaches typically require users to specify lower-level SLOs directly to guide resource allocation, creating a barrier for users without expertise in system configurations.

Sebrecht et al.~\cite{sebrecht} propose a fog mesh that accepts structured intents and parses them into workflows while considering user locality and workflow constraints to ensure QoS. However, the system does not support natural-language intent expression, maintaining an entry barrier for non-expert users.

Spillner et al.~\cite{spillner} explore microservice control in the compute continuum using high-level business objectives. Their approach adjusts resources of already deployed services to maintain performance targets but does not determine or recommend service placement based on user intent.

Morichetta et al.~\cite{schahram} investigate load balancing, cost efficiency, and function coordination in serverless environments using stakeholder-defined intents. Although the work references language-model-based translation, it ultimately requires users to provide explicit SLOs rather than natural-language specifications. \new{Their later inCoord framework~\cite{morichetta2025incoord} further shows that application-level intents in the cloud-edge continuum can be decomposed into domain-specific objectives across compute, network, and storage domains.}

Filinis et al.~\cite{filinis2024intent} present an intent-driven orchestration framework where users supply intent descriptions through keywords and metadata such as constraints and objectives. This reliance on structured parameters and low-level configurations limits accessibility for less-experienced users.

Akbari et al.~\cite{akbari2024icontinuum} introduce \emph{iContinuum}, an intent-driven emulation toolkit that provides a realistic testbed to evaluate placement strategies under user-defined metrics such as latency, privacy, and energy consumption. The framework focuses on benchmarking placement mechanisms rather than translating natural-language intents into actionable specifications.

Metsch et al.~\cite{metsch2023intent} propose an orchestration architecture for cloud-native deployments in which users declare objectives through structured Kubernetes Custom Resource Definitions specifying explicit Key Performance Indicators (KPIs) and SLO targets. A planner then translates these formal specifications into scaling and tuning actions. While effective for enforcing predefined objectives, the approach assumes that users already provide precise metric-level inputs and does not derive specifications from unstructured natural-language intents.
\new{
Sedlak et al.~\cite{sedlak2024diffusing} diffuse formal high-level SLOs in microservice pipelines into lower-level SLOs and parameter assignments using Bayesian networks learned from runtime metrics. However, it accepts formalized goals as SLOs and therefore does not address the translation of validated placement objectives from unstructured natural-language intents.
}

\subsection{Natural Language–Based Intent Translation}

To reduce the expertise required for low-level system orchestration, recent studies have explored natural-language intent interfaces that translate human-readable intents into formal SLO specifications consumable by traditional IDO frameworks~\cite{sebrecht, spillner, schahram, filinis2024intent, akbari2024icontinuum, metsch2023intent}.

Jacobs et al.~\cite{jacobs} introduce \emph{LUMI}, a chatbot-based interface that converts natural-language network intents into low-level configurations using a traditional named entity recognition pipeline based on word embeddings, Bi-LSTM models, and CRF tagging. While effective for short and structured utterances, the approach relies on predefined vocabularies and syntactic extraction, limiting its ability to handle complex or implicit intents and to generalize across heterogeneous environments.

Capova et al.~\cite{capova2025intent} translate natural-language business intents into reinforcement learning (RL) environments through knowledge-graph retrieval and LLM-based reasoning, generating reward functions and action spaces to guide RL agents. The translation produces learning objectives rather than SLO specifications that can be directly consumed by placement frameworks, and correctness depends on training convergence without deterministic or schema-constrained guarantees.

Esashi et al.~\cite{esashi2026action} propose \emph{Action Engine}, an LLM-driven framework that converts natural-language workflow descriptions into executable FaaS workflows by selecting functions and inferring dependencies to construct a workflow DAG. The method targets application-level workflow composition and synthesis, but does not derive formal system-level specifications or placement constraints required for orchestration control.

Asif et al.~\cite{asif2025evaluating} evaluate LLMs for natural-language intent translation in IBN policies and augment the pipeline with a KNN-based classifier to detect contradictory outputs. Since configurations are generated directly by the LLM and validated only post hoc, the process remains largely generative and lacks structured, deterministic guarantees during specification construction.

\new{
Mekrache et al.~\cite{mekrache2024llm} translate natural-language network management intents into Network Service Descriptors (NSDs) using KB-assisted few-shot prompting, structural validation, and human feedback. Mekrache et al.~\cite{mekrache2024intent} broaden this direction into an LLM-centric intent life-cycle architecture covering decomposition, translation, negotiation, activation, and assurance. These works target NSD generation and network intent life-cycle management, but they do not address placement-specific SLO construction where metric--operator--value constraints must be grounded against fluctuating edge--cloud resource states.

OSS-GPT~\cite{mekrache2025oss,mekrache2025next} uses assistant, planner, executor, and reporter agents to translate natural-language OSS intents into executable API-call sequences. DMO-GPT~\cite{mekrache2025dmo} extends this agentic design to distributed multi-operator 6G management by selecting operators and coordinating API execution across heterogeneous OSSs. These systems focus on direct OSS API planning and execution, where the generated output is tied to API payloads and service descriptors. As a result, they do not provide a standalone validated SLO abstraction that can be inspected, rejected, or reused by downstream placement engines before orchestration actions are executed.
}
Overall, these works demonstrate that natural-language interfaces can lower the expertise barrier for intent expression across domains. However, they largely rely on generative translation or post-hoc checks and do not incorporate contextual grounding or structured validation mechanisms necessary for reliable natural-language-to-SLO construction in dynamic compute-continuum environments.

\subsection{Service Placement Methods}

To optimize microservice placement across the compute continuum, placement algorithms are widely adopted, primarily focusing on compute and network resources~\cite{zafeiro}.

Samani et al.~\cite{samani} propose a coordination-based method to prevent QoS violations through Service-level Agreements (SLAs). Their approach considers both the current capabilities and historical credibility of devices when determining placement. However, Proactive Application Placement does not incorporate user intent, instead prioritizing SLA requirements defined by the infrastructure.

Mota-Cruz et al.~\cite{mota} present two approaches to minimize latency and balance load, termed \emph{app-based} and \emph{service-based} placement. The app-based method deploys all services sequentially within an application, whereas the service-based method evaluates each service independently while accounting for interconnected loads. Although the study compares different strategies based on application context and optimization objectives, it does not consider high-level user intent to guide placement decisions in the compute continuum.

\subsection{Retrieval-Augmented Grounding for LLM-Based Systems}

LLM-based intent translation can be improved either through model adaptation or inference-time grounding. While large-scale pretraining and domain-specific fine-tuning improve specialization, they are costly to maintain in rapidly changing system environments.

Inference-time methods condition the model without modifying its parameters. Zero-shot prompting~\cite{kojima2022large} and few-shot in-context learning~\cite{wei2022chain} provide lightweight adaptation, while retrieval-augmented generation (RAG) grounds responses using external context~\cite{lewis2020retrieval}. This is particularly relevant for distributed-system intent translation, where implicit requirements often depend on fluctuating runtime metrics; retrieving live or historical continuum state enables concrete value resolution instead of prompt-only estimation.

Several recent approaches use retrieval and reasoning for executable task generation. Zhang et al.~\cite{zhang2024reverse} propose \emph{Reverse Chain}, which decomposes requests into API selection and argument completion. Yao et al.~\cite{yao2022react} introduce \emph{ReAct}, which interleaves reasoning with environment actions to retrieve evidence during decision making. Verma et al.~\cite{verma2024plan} propose \emph{Plan-RAG}, which decomposes queries into a DAG of sub-tasks for targeted retrieval and generation.

These approaches improve grounded reasoning and reduce hallucination, but they primarily target workflow planning, API generation, or question answering. They do not directly construct formally constrained system-level specifications for orchestration, where retrieved evidence must be combined with schema validation and control-plane compatibility.

\subsection{Research Gaps}

Existing IDO frameworks typically require users to provide explicit SLO specifications, shifting complexity to users who need domain expertise and contextual system knowledge. Natural-language intent interfaces reduce this barrier, \new{but existing approaches remain limited for compute-continuum placement because they often depend on direct prompt-based generation or post-hoc validation. Such approaches provide limited support for intents whose correct specification depends on monitored infrastructure state, such as ``highest bandwidth'' or ``lowest memory utilization''.}

\new{A gap therefore remains in reliably transforming natural-language placement intents into orchestration-consumable SLO artifacts before they are used by downstream placement logic. This requires not only language interpretation, but also schema-bounded specification construction, infrastructure-aware grounding, and rejection of unsupported, ambiguous, or conflicting requests. This paper addresses this gap through \emph{Intent Engine}.}

\section{System Architecture}
\label{system_architecture}

This section presents the proposed \emph{Intent Engine} architecture as a control-plane SLO construction system for the compute continuum. It is designed to integrate with existing intent-driven orchestration (IDO) and placement frameworks without replacing their placement algorithms or runtime assurance mechanisms.

\new{Reliable SLO construction is critical because the translated output is intended to become an orchestration-consumable control-plane artifact for a downstream IDO framework. If an intent is mistranslated, the downstream framework may receive unsupported constraints, incorrect resource values, conflicting objectives, or malformed specifications. Such errors can lead to unintended service placement, SLO violations, inefficient resource use, or application disruption once the specification is consumed by an orchestrator. Therefore, \emph{Intent Engine} treats intent translation as a controlled specification-construction process rather than unconstrained end-to-end text generation.}

\begin{figure*}[t]
  \centering
  \includegraphics[width=1\textwidth]{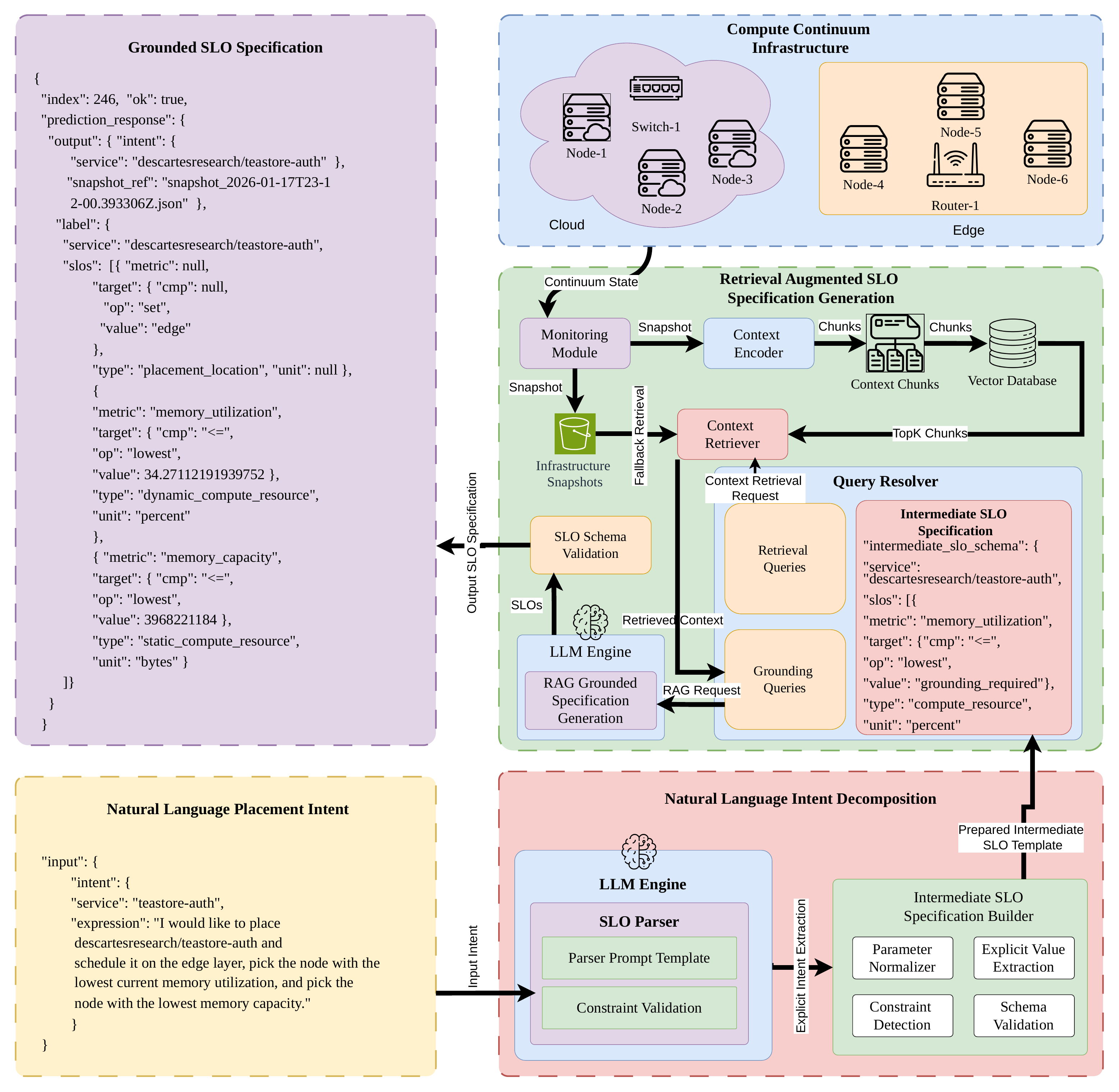}
  \caption{Overview of the proposed \emph{Intent Engine} system architecture.}
  \label{figure:system_architecture}
\end{figure*}

Figure~\ref{figure:system_architecture} illustrates the proposed architecture. The pipeline first extracts semantic constraints from the natural-language intent into an intermediate SLO representation (IR). It then grounds implicit or context-dependent requirements by retrieving evidence from the compute-continuum infrastructure state. Finally, structural and schema validation is applied before producing the final SLO specification, ensuring compatibility with existing orchestration and placement engines. \new{\emph{Intent Engine} is limited to intent acquisition and SLO construction; placement decisions, deployment execution, runtime intent assurance, and re-grounding are handled by the downstream IDO framework that consumes the generated specification.}

The architecture consists of two primary components: (i) \emph{Natural Language Intent Decomposition} and (ii) \emph{Retrieval-Augmented SLO Specification Generation}. The former extracts explicit constraints, identifies implicit placement requirements, and prepares an intermediate SLO representation. The latter grounds implicit requirements through contextual system-state retrieval and produces the final validated SLO specification. Together, these components form the artifact boundary between user intent and downstream orchestration logic: only schema-valid and infrastructure-grounded SLO specifications are emitted. The core modules are described below.

\subsection{System Input}

The architecture accepts two inputs for each placement request: the \emph{service name} and a natural-language \emph{intent expression}. The intent may describe explicit constraints (e.g., ``memory utilization below 60\%''), implicit preferences (e.g., ``highest memory utilization''), or deployment location requirements (e.g., ``deploy in the cloud'').

To ensure compatibility with existing IDO frameworks~\cite{sebrecht, spillner, schahram, filinis2024intent, akbari2024icontinuum, metsch2023intent}, we define a supported SLO schema shown in Table~\ref{table:slo_constraints_table}. \new{This infrastructure-oriented schema is consistent with inCoord by Morichetta et al., where application-level intents are decomposed into domain-specific compute, network, and storage objectives. The selected constraint types reflect the acquisition specifications commonly used by existing IDO and placement frameworks, including placement location, compute capacity, utilization, and network-related constraints. The schema is also configurable, allowing framework-specific metrics, units, and constraint types to be added when required. Schema validation checks whether each generated SLO uses only supported metrics, operators, units, and value formats before the specification is emitted. Restricting translation to this predefined schema preserves downstream compatibility, limits the LLM output space, and reduces unsupported or hallucinated constraints.} This design enables \emph{Intent Engine} to operate as a flexible front-end intent acquisition component for existing frameworks.

\subsection{Natural Language Intent Decomposition}

The intent decomposition component transforms an unstructured natural-language intent into a structured, schema-compliant intermediate SLO representation. It consists of two modules: the \emph{SLO Parser} and the \emph{Intermediate SLO Specification Builder}. Unlike prior approaches that directly generate final configuration or policy artifacts~\cite{capova2025intent, asif2025evaluating}, our architecture separates intent understanding, constraint extraction, and schema construction before contextual grounding.

\subsubsection{SLO Parser}

The SLO Parser extracts constraints from the service-specific natural-language intent using an LLM guided by the parser prompt template shown in Figure~\ref{figure:parser_prompt_template}. The prompt directs the model to identify explicitly stated metrics, comparison operators, and values while adhering to the supported SLO schema and normalized unit formats.

This stage focuses on information explicitly present in the intent and prepares placeholders for requirements that need contextual grounding. The prompt enforces: 
(i) adherence to predefined SLO categories (Table~\ref{table:slo_constraints_table}), 
(ii) canonical unit and operator normalization, 
(iii) a fixed intermediate SLO schema, and 
(iv) exclusion of unsupported parameters.

\new{Because the IR is extracted by an LLM, it may still include unsupported capabilities, malformed units or operators, or conflicting constraints. To prevent these errors from propagating, the SLO Parser applies \emph{Constraint Validation} using Algorithm~\ref{alg:llm_ir_validation}. The algorithm checks the generated IR against the supported capability registry and rejects invalid or inconsistent clauses before grounded SLO construction, preventing unsafe intermediate outputs from reaching downstream orchestration logic.}

\subsubsection{Intermediate SLO Specification Builder}

The \emph{Intermediate SLO Specification Builder} post-processes the parser output into a schema-compliant intermediate representation for contextual grounding. It normalizes metrics, operators, and units; maps explicit values to canonical fields; verifies conformity with the supported SLO schema; and marks constraints without explicit values as requiring grounding.

This intermediate representation acts as a control point between language understanding and final specification generation. By decomposing the intent into validated components, the architecture avoids monolithic LLM generation and constructs the final specification through guided, context-aware reasoning.

\subsection{Retrieval-Augmented SLO Specification Generation}

This component takes the intermediate SLO representation and grounds implicit requirements by retrieving contextual information from the compute continuum. It addresses cases where prompt-only generation is insufficient, particularly extrema-based requests such as ``highest memory'' or constraints that depend on fluctuating system conditions.

\subsubsection{Query Resolver}

The Query Resolver identifies constraints that require contextual grounding and formulates retrieval queries specifying the metric, scope, and grounding objective, such as highest or lowest resource value. It then combines the intermediate SLO template with retrieved evidence to guide grounded value generation while preserving the predefined schema.

\subsubsection{Context Retriever and Encoder}

The Context Retriever gathers compute-continuum information from two sources: (i) a vector database containing embedded system-state representations and (ii) a fallback snapshot stored in file storage. Retrieval first performs a top-$k$ similarity search on the vector database\footnote{$k=8$ in our implementation.}; if the retrieved evidence is insufficient or unavailable, the system falls back to snapshot-based context retrieval.

The Context Encoder converts infrastructure snapshots into retrieval-ready chunks containing node-level states, aggregated edge--cloud summaries, and structured resource metadata. The encoding process is deterministic and does not rely on an LLM, ensuring consistent context representation and avoiding additional generative errors in the grounding source.

\subsubsection{Monitoring Module}

The monitoring module continuously collects static and dynamic resource metrics across edge and cloud nodes, including compute capacities, utilization levels, and network statistics. Metrics are sampled periodically\footnote{every 10 minutes in our implementation.} and stored as both raw snapshots and encoded representations for retrieval. This module provides the infrastructure state required for reliable grounding.

\subsection{Compute Continuum Infrastructure}

The compute continuum infrastructure comprises heterogeneous edge and cloud resources managed by existing orchestration frameworks. In this work, the infrastructure primarily supplies the contextual state required for grounding placement intents and constructing ground-truth SLO labels. The proposed architecture is agnostic to the underlying infrastructure implementation, enabling integration with a wide range of edge--cloud platforms and orchestration systems. Evaluating the placement decisions made after an IDO framework consumes the generated SLOs is outside the scope of the translation layer studied here.

\section{Implementation and Experimental Setup}
\label{implementation_and_experimental_setup}

We implemented \new{\emph{Intent Engine} as an intent-to-SLO construction layer and evaluated its accuracy, robustness, and grounding effectiveness}. Experiments were conducted on a real compute-continuum testbed, enabling systematic comparison with LLM-based intent translation baselines under realistic edge--cloud conditions.

The implementation consists of two main components: (i) a compute continuum testbed spanning edge and cloud layers, and (ii) the \emph{Intent Engine} pipeline for intent decomposition, retrieval-augmented grounding, and SLO specification generation. The testbed supplies periodically monitored system context for grounding implicit constraints and deriving labeled SLO targets, while the pipeline executes the SLO construction process described in Section~\ref{system_architecture}. The testbed is not used to evaluate downstream placement quality or runtime QoS optimization, which depend on the IDO framework that consumes the generated SLOs. Each component is described below. 

\subsection{Compute Continuum Infrastructure Testbed}

We built a physical compute continuum testbed spanning geographically distributed edge and cloud resources to collect realistic resource-state traces under deployment-like conditions and resource variability. Since retrieval-augmented grounding depends on current system context, the infrastructure continuously collects resource-state data used by the SLO grounding pipeline and by the construction of ground-truth labels.

\begin{figure*}[t]
  \centering
  \includegraphics[width=1\textwidth]{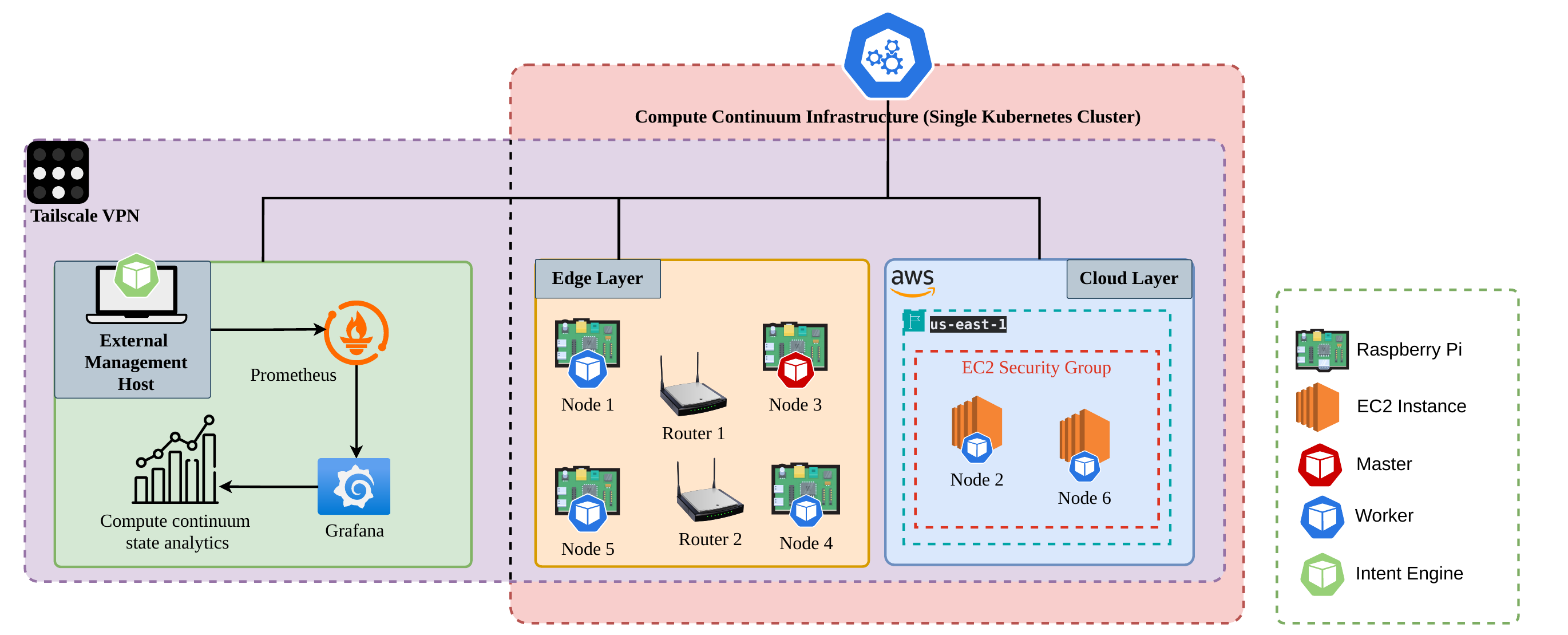}
  \caption{Overview of the real compute continuum testbed and resource monitoring infrastructure.}
  \label{figure:compute_continuum_testbed}
\end{figure*}

The testbed comprises heterogeneous compute and network resources interconnected across the edge--cloud continuum, as shown in Figure~\ref{figure:compute_continuum_testbed}. Table~\ref{table:compute_resources} summarizes the hardware specifications of the deployed nodes.

\begin{table*}[t]
\centering
\caption{Compute resource specification of the compute continuum testbed.}
\label{table:compute_resources}
\begin{tabular}{l p{3.2cm} l l l l l} \hline
Node & Model & CPU & Memory & Storage & Role & Location \\ \hline
Node~1 & Raspberry~Pi~4      & 2 cores & 4\,GB & 64\,GB  & Worker & Edge \\
Node~2 & AWS EC2 Instance    & 2 cores & 2\,GB & 128\,GB & Worker & Cloud \\
Node~3 & Raspberry~Pi~5      & 4 cores & 8\,GB & 64\,GB  & Master & Edge \\
Node~4 & Raspberry~Pi~4      & 4 cores & 4\,GB & 64\,GB  & Worker & Edge \\
Node~5 & Raspberry~Pi~3      & 2 cores & 4\,GB & 32\,GB  & Worker & Edge \\
Node~6 & AWS EC2 Instance    & 2 cores & 2\,GB & 80\,GB  & Worker & Cloud \\
\hline
\end{tabular}
\end{table*}

All six nodes are managed as a single Kubernetes~\cite{kubernetes_docs} cluster spanning edge and cloud locations. Node~3 operates as the master node responsible for cluster orchestration and scheduling, while the remaining nodes act as workers interconnected through Router~1 and Router~2 to expose real network variability.

The edge layer consists of heterogeneous Raspberry Pi devices, providing diverse resource capacities for trace collection and grounding. The cloud layer is provisioned using Amazon Web Services (AWS) in the \emph{us-east-1} region, where EC2~\cite{ec2_docs} instances are configured within a dedicated \emph{EC2 Security Group} and integrated as Kubernetes worker nodes. Although AWS is used in our deployment, the architecture remains cloud-provider agnostic.

\new{An External Management Host is used for Kubernetes cluster management, infrastructure control, and visualization of continuum-state analytics, supporting the implementation of the \emph{Monitoring Module}. This host is external to the compute continuum; service placement and orchestration are executed within the cluster by the master node. \emph{Intent Engine} runs on the External Management Host, and its internal architecture includes the LLM engine, intent decomposition, context retrieval, and retrieval-grounded SLO generation components.}

Network connectivity spans distinct access routers to emulate distributed domains. A secure virtual private network (VPN) overlay is established using Tailscale~\cite{tailscale_docs} to enable private communication between edge and cloud nodes. To emulate realistic deployment scenarios, we deploy the TeaStore~\cite{teastore} reference microservice application, comprising six interacting services, using Kubernetes-native deployment.

\subsection{Natural Language Intent Decomposition}

In our implementation, service placement intents are supplied as unstructured natural-language inputs and processed by the intent decomposition component to produce structured intermediate specifications.

\subsubsection{SLO Parser}

The \emph{SLO Parser} is implemented using an LLM engine guided by the parser prompt template shown in Figure~\ref{figure:parser_prompt_template}. We use GPT-4.1-mini through the OpenAI~\cite{openai_docs} API as the underlying model. The temperature is set to 0.0 to preserve extraction stability and reduce unnecessary output variation.

The parser is implemented in Python and exposed as a lightweight Flask~\cite{flask_docs}-based API service for integration with the orchestration pipeline. The API accepts JavaScript Object Notation (JSON) requests containing the service identifier and intent text, and returns a structured JSON response that is forwarded to subsequent stages for normalization and grounding.

\new{
\begin{table*}[t]
\caption{Supported constraint space used by the generated IR and downstream grounding. Each capability $c$ is associated with an admissible domain $\mathcal{D}(c)$ and a canonical representation $\bar{d}(c)$.}
\label{table:slo_constraints_table}
\centering
\begin{tabular}{l|l|l|l}
\hline
\multirow{2}{*}{Category}
& \multirow{2}{*}{Capability $c$}
& \multirow{2}{*}{Admissible Domain $\mathcal{D}(c)$}
& \multirow{2}{*}{Canonical Representation $\bar{d}(c)$} \\
& & & \\
\hline

Placement
& Placement Location
& Edge, Cloud, Both, Any
& Edge / Cloud / Both / Any \\
\hline

\multirow{3}{*}{Static Compute}
& CPU Capacity
& Cores, Millicores
& Cores \\
\cline{2-4}
& Memory Capacity
& Bytes, B, KB, MB, GB, KiB, MiB, GiB
& Bytes \\
\cline{2-4}
& Storage Capacity
& Bytes, B, KB, MB, GB, KiB, MiB, GiB
& Bytes \\
\hline

\multirow{3}{*}{Dynamic Compute}
& CPU Utilization
& \%
& Percent \\
\cline{2-4}
& Memory Utilization
& \%
& Percent \\
\cline{2-4}
& Storage Utilization
& \%
& Percent \\
\hline

\multirow{2}{*}{Node Network}
& Bandwidth
& Bps, Kbps, Mbps, Gbps
& Bps \\
\cline{2-4}
& Port Utilization
& \%
& Percent \\
\hline
\end{tabular}
\end{table*}
}

\subsubsection{Intermediate SLO Specification Builder}

This module post-processes the parser output to construct a schema-compliant intermediate SLO representation. It enforces the predefined schema by normalizing field names, inserting missing fields with null values, and validating placement and parameter constraints. It also canonicalizes metric names, units, and operators, and annotates each constraint with a \texttt{grounding\_required} flag to indicate whether contextual grounding is needed.

\new{
\begin{algorithm}[t]
\footnotesize
\caption{LLM-Generated IR Validation}
\label{alg:llm_ir_validation}
\begin{algorithmic}[1]
\Require LLM-generated IR $I$, supported capability registry $\mathcal{R}$
\Ensure validated intent IR $I_{\mathrm{valid}}$

\State $I_{\mathrm{valid}} \gets [\,]$, \quad $p \gets \texttt{Any}$
\State initialize $lb[c] \gets -\infty$, $ub[c] \gets +\infty$, $ext[c] \gets \varnothing$ for each numeric capability $c \in \mathcal{R}$
\Comment{$lb[c],ub[c]$: feasible bounds; $ext[c]$: extremum request}

\ForAll{generated clause $s \in I$}
    \State $(c,r,v) \gets \textsc{Normalize}(s,\mathcal{R})$
    \Comment{$c$: capability; $r$: relation; $v \in \bar{d}(c)$}

    \If{$(c,r,v)$ is invalid}
        \State \Return $\emptyset$
        \Comment{unsupported capability or invalid unit/value}
    \EndIf

    \Statex \textit{Non-numeric validation}
    \If{$c = \texttt{Placement Location}$}
        \If{$v = \texttt{Any}$}
            \State append $s$ to $I_{\mathrm{valid}}$; \textbf{continue}
            \Comment{no placement restriction}
        \ElsIf{$p = \texttt{Any}$ or $p = v$}
            \State $p \gets v$
        \ElsIf{$p = \texttt{Both}$ or $v = \texttt{Both}$}
            \State $p \gets \texttt{Both}$
        \ElsIf{$(p=\texttt{Edge} \land v=\texttt{Cloud})$ or $(p=\texttt{Cloud} \land v=\texttt{Edge})$}
            \State $p \gets \texttt{Both}$
            \Comment{merge Edge and Cloud}
        \Else
            \State \Return $\emptyset$
            \Comment{conflicting placement values}
        \EndIf
        \State append $s$ to $I_{\mathrm{valid}}$; \textbf{continue}
    \EndIf

    \Statex \textit{Numeric validation}
    \If{$r = \texttt{>=}$ or $r = \texttt{=}$}
        \State $lb[c] \gets \max(lb[c], v)$
    \EndIf
    \If{$r = \texttt{<=}$ or $r = \texttt{=}$}
        \State $ub[c] \gets \min(ub[c], v)$
    \EndIf

    \If{$lb[c] > ub[c]$}
        \State \Return $\emptyset$
        \Comment{empty feasible interval}
    \EndIf

    \If{$r = \texttt{max}$}
        \If{$ext[c] = \texttt{min}$}
            \State \Return $\emptyset$
            \Comment{opposite extrema}
        \EndIf
        \State $ext[c] \gets \texttt{max}$
    \ElsIf{$r = \texttt{min}$}
        \If{$ext[c] = \texttt{max}$}
            \State \Return $\emptyset$
            \Comment{opposite extrema}
        \EndIf
        \State $ext[c] \gets \texttt{min}$
    \EndIf

    \State append $s$ to $I_{\mathrm{valid}}$
\EndFor

\State \Return $I_{\mathrm{valid}}$
\Comment{forward only supported and consistent IR}
\end{algorithmic}
\end{algorithm}
}

The resulting intermediate specification is serialized as a JSON artifact and forwarded to the retrieval-augmented grounding stage.

\subsection{Retrieval-Augmented SLO Specification Generation}

The retrieval-augmented generation (RAG) component grounds intent constraints using monitored compute continuum context. It accepts the intermediate specification as a JSON artifact, detects parameters requiring grounding, retrieves contextual information, generates final SLO specifications, and validates structural and schema completeness.

\subsubsection{Monitoring Module}

We monitor the compute continuum testbed using a kube-prometheus stack. Prometheus~\cite{prometheus_docs} collects static resource capacities and dynamic utilization metrics, including CPU, memory, storage, and network statistics.

Grafana~\cite{grafana_docs} is used for visualization and manual inspection of resource trends. Additionally, we implement a Python Flask~\cite{flask_docs}-based API service that periodically queries Prometheus through node-exporter every 10 minutes, extracts point-in-time infrastructure snapshots, and stores them as JSON artifacts in Amazon S3~\cite{s3_docs}. These snapshots serve as a deterministic fallback source for contextual grounding.

Each snapshot is also forwarded to the \emph{Context Encoder} for embedding and storage in the vector database.

\subsubsection{Context Encoder}

The Context Encoder processes each infrastructure snapshot into retrieval-ready context chunks. Each chunk captures node-level state, network measurements, or aggregated summaries, and is encoded as embedding-friendly text paired with structured metadata. The chunks are stored in a Qdrant~\cite{qdrant_docs} vector database to support similarity-based retrieval.

Chunking is implemented deterministically rather than through LLM-based generation to ensure reproducibility and avoid introducing spurious context. During retrieval, only the top-$k$ most relevant chunks are returned; in our implementation, $k=8$.

\subsubsection{Context Retriever}

The Context Retriever resolves metric queries using a two-stage strategy. Queries are first executed against the Qdrant vector database using similarity search to obtain relevant chunks. If retrieval is insufficient or unavailable, the system falls back to the most recent snapshot stored in Amazon S3~\cite{s3_docs}.

Retrieved values are then aggregated deterministically, with unit normalization and metadata filtering applied as required.

\subsubsection{Query Resolver}

The Query Resolver analyzes the intermediate SLO specification and identifies parameters requiring contextual grounding. For each parameter, it generates structured retrieval queries that encode the desired aggregation semantics, such as highest or lowest.

Retrieved context is combined with grounding prompts and passed to the LLM engine to instantiate concrete values. Explicitly specified constraints bypass retrieval and are propagated directly. The grounded outputs are then canonicalized to a fixed schema, with metric names, units, and operators normalized to match the conventions used in the infrastructure snapshots, ensuring consistent grounding and stable specification construction.

\subsubsection{SLO Specification Validation}

The final SLO specification is validated for structural completeness, schema consistency, and value correctness. Units and field names are checked against the snapshot representations to guarantee alignment with the monitored system state. The validated specification is serialized as a JSON artifact and forms the final system output.

\section{Evaluation}
\label{evaluation}
\new{
This section evaluates whether \emph{Intent Engine} can reliably construct accurate SLO artifacts from natural-language placement intents for downstream orchestration. The evaluation focuses on the intent acquisition and SLO construction layer rather than runtime orchestration behavior. Specifically, it examines: 
(i) whether generated SLO constraints match the ground truth, 
(ii) whether the complete specification is structurally correct, 
(iii) whether hallucinations are reduced, 
(iv) whether retrieval grounding improves implicit value resolution, 
(v) whether invalid intents are safely rejected, 
(vi) whether the construction pipeline operates within acceptable latency and scalability limits, and
(vii) how translation errors affect downstream placement feasibility.
}
\subsection{Dataset}
\label{evaluation_dataset}

To the best of our knowledge, no public dataset exists for grounded natural-language intent-to-SLO translation in a compute-continuum environment. \new{We therefore construct a dataset from our real testbed and augment it with synthetic natural-language variants to increase linguistic diversity while preserving the target SLO schema}

\new{Table~\ref{table:dataset_complexity_levels} depicts the dataset containing 716 intent records across five complexity levels. Of these, 521 records are valid intent-to-SLO pairs, and 195 records are invalid cases covering ambiguous, conflicting, malformed, and unsupported intents. Valid records are used for translation-correctness evaluation, while invalid records support robustness and failure-handling analysis. Each record is associated with a point-in-time compute-continuum snapshot used to derive the ground-truth labels and support contextual grounding during evaluation.}

\begin{table*}[!t]
\centering
\caption{Dataset composition by intent complexity level, source type, validity status, and number of SLOs per valid intent.}
\label{table:dataset_complexity_levels}

\small
\setlength{\tabcolsep}{4.5pt}
\renewcommand{\arraystretch}{1.15}

\begin{tabular*}{\textwidth}{@{\extracolsep{\fill}}lcccccc@{}}
\hline
\textbf{Complexity Level}
& \textbf{Total}
& \textbf{Real}
& \textbf{Synthetic}
& \textbf{Valid}
& \textbf{Invalid}
& \textbf{SLOs per Valid Intent} \\
\hline
Level 1 & 189 & 59 & 130 & 137 & 52 & 1 \\
Level 2 & 207 & 67 & 140 & 174 & 33 & 2--3 \\
Level 3 & 177 & 57 & 120 & 141 & 36 & 2--3 \\
Level 4 & 93  & 23 & 70  & 48  & 45 & 3--4 \\
Level 5 & 50  & 10 & 40  & 21  & 29 & 2--3 \\
\hline
\textbf{Total} 
& \textbf{716} 
& \textbf{216} 
& \textbf{500} 
& \textbf{521} 
& \textbf{195} 
& -- \\
\hline
\end{tabular*}
\end{table*}

\new{Real records are derived from the physical edge--cloud testbed using monitored infrastructure snapshots. Synthetic records are generated using the Llama 3.3 70B model to expand the linguistic diversity of intents associated with the real testbed traces. This augmentation is used only to increase natural-language variation while preserving the same schema, supported metrics, and complexity structure. To avoid synthetic-label noise, all records are checked against the supported SLO schema, and implicit values are derived from the corresponding snapshots rather than accepted from the generation model. Invalid records are retained rather than discarded so that the evaluation also captures ambiguous, conflicting, malformed, and unsupported user intents.

The TeaStore microservice application is used only as a reference workload to generate realistic service identifiers, placement scenarios, and infrastructure traces. The proposed architecture is application-agnostic: \emph{Intent Engine} constructs SLO artifacts over the supported schema and monitored infrastructure state, regardless of the specific application used to produce the traces. TeaStore therefore provides a concrete service context for dataset construction, but the evaluated task is not TeaStore-specific.

Complexity levels reflect the number and type of constraints in an intent. Level~1 contains a single SLO. Levels~2--4 contain increasing combinations of placement, compute, and network constraints. Level~5 contains the most challenging cases, where implicit snapshot-grounded requirements are often combined with explicit placement or threshold constraints. Dataset examples, including valid and invalid records across complexity levels, are illustrated in Figure~\ref{figure:dataset_sample} in the Appendix.}

\subsection{Candidate Models}
\label{candidate_models}
\new{
Table~\ref{table:candidate_models} summarizes the LLMs used in dataset construction and evaluation. Llama 3.3 70B is used only for synthetic intent generation, while GPT-4.1 mini, Claude Sonnet 4.5, and DeepSeek V4-Flash are used as translation evaluation backends. This separation reduces model-bias risk because the model used to diversify the synthetic intents is not used to evaluate the architecture. Using multiple closed- and open-source evaluation backends further tests whether the observed gains generalize across model families rather than depending on a single LLM.

\begin{table}[!t]
\centering
\caption{Candidate LLMs used for synthetic dataset generation and translation evaluation.}
\label{table:candidate_models}

\footnotesize
\setlength{\tabcolsep}{3.5pt}
\renewcommand{\arraystretch}{1.15}

\begin{tabular}{p{0.30\columnwidth}p{0.13\columnwidth}p{0.48\columnwidth}}
\hline
\textbf{Model}
& \textbf{Access}
& \textbf{Role} \\
\hline
Llama 3.3 70B
& Open
& Synthetic intent generation. \\

GPT-4.1 mini
& Closed
& Translation evaluation backend. \\

Claude Sonnet 4.5
& Closed
& Translation evaluation backend. \\

DeepSeek V4-Flash
& Open
& Translation evaluation backend. \\
\hline
\end{tabular}
\end{table}
}

\subsection{Baseline Methods}
\label{baseline_methods}

To the best of our knowledge, no prior system directly translates unstructured natural-language placement intents into retrieval-grounded orchestration-consumable SLO specifications for compute-continuum environments. We therefore compare \emph{Intent Engine} against four prompt-only LLM baselines commonly used in recent natural-language intent translation work~\cite{capova2025intent, esashi2026action}: \emph{Zero-shot}, \emph{Few-shot}, \emph{Zero-shot Chain-of-Thought (CoT)}, and \emph{Few-shot CoT}. Task-specific pretrained or fine-tuned baselines are not included because no established labeled dataset or pretrained model exists for this domain-specific grounded intent-to-SLO translation task.

\new{To include a non-prompting structured alternative, we implement a \emph{Rule-based Parser} baseline. It uses schema constrained metric, operator, placement, and unit rules with snapshot resolution for implicit highest/lowest constraints. The parser uses no prompts, LLM outputs, learned parameters, or dataset-specific templates, making it a generic non-LLM baseline under the same SLO schema.}

\new{For evaluation fairness, the prompting baselines are also provided with the infrastructure snapshots when grounded values are required. This gives the baselines access to the same contextual evidence, although it increases prompt length and complexity compared with the retrieval and schema-bounded pipeline used by \emph{Intent Engine}. This baseline setting differs from \emph{Intent Engine}, which retrieves only metric- and intent-relevant infrastructure context instead of injecting the full continuum snapshot into a single prompt. This reduces unnecessary context length and avoids exposing the LLM to unrelated node-state information.}

\new{We evaluate the baselines and \emph{Intent Engine} using three backend LLMs: GPT-4.1 mini, Claude Sonnet 4.5, and DeepSeek V4-Flash.} The same prompt design is applied consistently across backends. The prompt templates for the baseline methods are shown in Figure~\ref{figure:zero_shot_and_cot_prompt_template} and Figure~\ref{figure:fews_shot_and_cot_prompt_template}. The ablation study uses GPT-4.1 mini to isolate the effect of retrieval grounding while holding the backend model fixed.

\subsection{Metrics and Definitions}
\label{notation_and_definitions}

Let $\mathcal{D}=\{(x_i, y_i)\}_{i=1}^{N}$ denote the dataset, where $x_i=(s_i,t_i)$ consists of a service identifier $s_i$ and a natural-language intent expression $t_i$, and $y_i$ is the corresponding ground-truth SLO specification. The dataset contains $N=716$ records, with a valid subset $\mathcal{D}_{\mathrm{valid}}$ of 521 valid intent-to-SLO pairs. Unless otherwise stated, translation-correctness metrics are computed on $\mathcal{D}_{\mathrm{valid}}$.

Each record $i$ is associated with a system-state snapshot $c_i$ captured from the testbed at labeling time. This snapshot is not provided as a user input; it is used to derive ground-truth labels and support contextual grounding during evaluation.

For each record $i$, we represent the ground-truth SLO specification as a set of atomic constraints $\mathcal{S}_i^{\mathrm{gt}}$ and the predicted specification as $\mathcal{S}_i^{\mathrm{pred}}$. Each atomic constraint is canonicalized as a tuple of resource type, metric, operator/comparator, value, and normalized unit. A predicted constraint is counted as a true positive (TP) when it exactly matches a ground-truth constraint under this canonical representation. A false positive (FP) is a predicted constraint with no matching ground-truth constraint, and a false negative (FN) is a ground-truth constraint missing from the prediction.

\new{We use strict matching rather than mean absolute error (MAE) or tolerance-based scoring because SLOs and SLAs are threshold-based control-plane artifacts that can be consumed by downstream placement or orchestration frameworks. A small numeric deviation can still change whether a constraint is satisfied or violated, and therefore can affect downstream placement decisions. Exact matching better reflects the reliability requirement of validated SLO construction.}

For a method $m$ and complexity level $l$, we aggregate true positives ($TP_{m,l}$), false positives ($FP_{m,l}$), and false negatives ($FN_{m,l}$), and compute:
\begin{align}
\text{Precision}_{m,l} &= \frac{TP_{m,l}}{TP_{m,l}+FP_{m,l}}, \label{eq:precision}\\
\text{Recall}_{m,l} &= \frac{TP_{m,l}}{TP_{m,l}+FN_{m,l}}, \label{eq:recall}\\
\text{F1}_{m,l} &= \frac{2\cdot \text{Precision}_{m,l}\cdot \text{Recall}_{m,l}}{\text{Precision}_{m,l}+\text{Recall}_{m,l}}. \label{eq:f1}
\end{align}

For specification-level structure, we use Exact Match and Jaccard similarity:
\begin{equation}
\text{ExactMatch}(i)=
\begin{cases}
1, & \text{if } \mathcal{S}_i^{\text{pred}}=\mathcal{S}_i^{\text{gt}},\\
0, & \text{otherwise},
\end{cases}
\label{eq:exactmatch}
\end{equation}
\begin{equation}
\text{Jaccard}(i)=\frac{\left|\mathcal{S}_i^{\text{pred}}\cap \mathcal{S}_i^{\text{gt}}\right|}{\left|\mathcal{S}_i^{\text{pred}}\cup \mathcal{S}_i^{\text{gt}}\right|}.
\label{eq:jaccard}
\end{equation}

To quantify hallucinations, we track three binary indicators per response:
\begin{align}
h^{(1)}_i &= \mathbbm{1}\{\text{unsupported constraint occurs in response } i\}, \nonumber\\
h^{(2)}_i &= \mathbbm{1}\{\text{incorrect or missing value in response } i\}, \nonumber\\
h^{(3)}_i &= \mathbbm{1}\{\text{schema violation occurs in response } i\}. \nonumber
\end{align}
Average rates are:
\begin{equation}
H_k = \frac{1}{N}\sum_{i=1}^{N} h^{(k)}_i,\quad k\in\{1,2,3\},
\label{eq:hallucination_components}
\end{equation}
and the aggregate hallucination score is
\begin{equation}
H = H_1 + H_2 + H_3,
\label{eq:hallucination_total}
\end{equation}
where $H$ is the average number of hallucination categories triggered per response. 

\new{
For invalid intents, the desired behavior is to reject the request or return no orchestration-consumable SLO specification. We therefore define the rejection rate as:
\begin{equation}
\mathrm{RejectionRate}_{m,c} =
\frac{N^{\mathrm{reject}}_{m,c}}{N^{\mathrm{invalid}}_{c}},
\label{eq:rejection_rate}
\end{equation}
where $N^{\mathrm{reject}}_{m,c}$ denotes the number of invalid intents in category $c$ correctly rejected by method $m$, and $N^{\mathrm{invalid}}_{c}$ denotes the total number of invalid intents in that category. Higher values indicate safer failure handling. A false accept occurs when an invalid intent is incorrectly converted into one or more validated SLO constraints.

For downstream placement-impact analysis, we use the deterministic placement generator as an evaluator. Let $N_{\mathrm{feasible}}$ denote the number of valid records whose ground-truth SLO label has at least one feasible placement in the corresponding snapshot. For a method $m$, let $N^{m}_{\mathrm{no\_match}}$ denote the number of cases where the predicted SLOs produce no matching node, and let $N^{m}_{\mathrm{invalid}}$ denote the number of cases where the predicted SLOs return a node that does not satisfy at least one constraint in the ground-truth SLO label. The placement failure rate is defined as:
\begin{equation}
\mathrm{PlacementFailure}_{m}
=
\frac{
N^{m}_{\mathrm{no\_match}} + N^{m}_{\mathrm{invalid}}
}{
N_{\mathrm{feasible}}
}.
\label{eq:placement_failure}
\end{equation}
}

\subsection{Hallucination Analysis}
\label{hallucination_analysis}

\new{We compute hallucination rates over the full dataset of $N=716$ records, including both valid and invalid intents. Correct rejection of an invalid intent does not trigger a hallucination category; if an invalid intent is accepted and converted into SLO constraints, the output is evaluated using the same canonical representation.} $H_1$ captures unsupported or spurious constraints, $H_2$ captures missing or incorrect values for otherwise matched constraints, and $H_3$ captures structural or schema violations such as invalid resource types, malformed operators, or inconsistent units.

\begin{figure*}[t]
  \centering
  \includegraphics[width=0.85\textwidth]{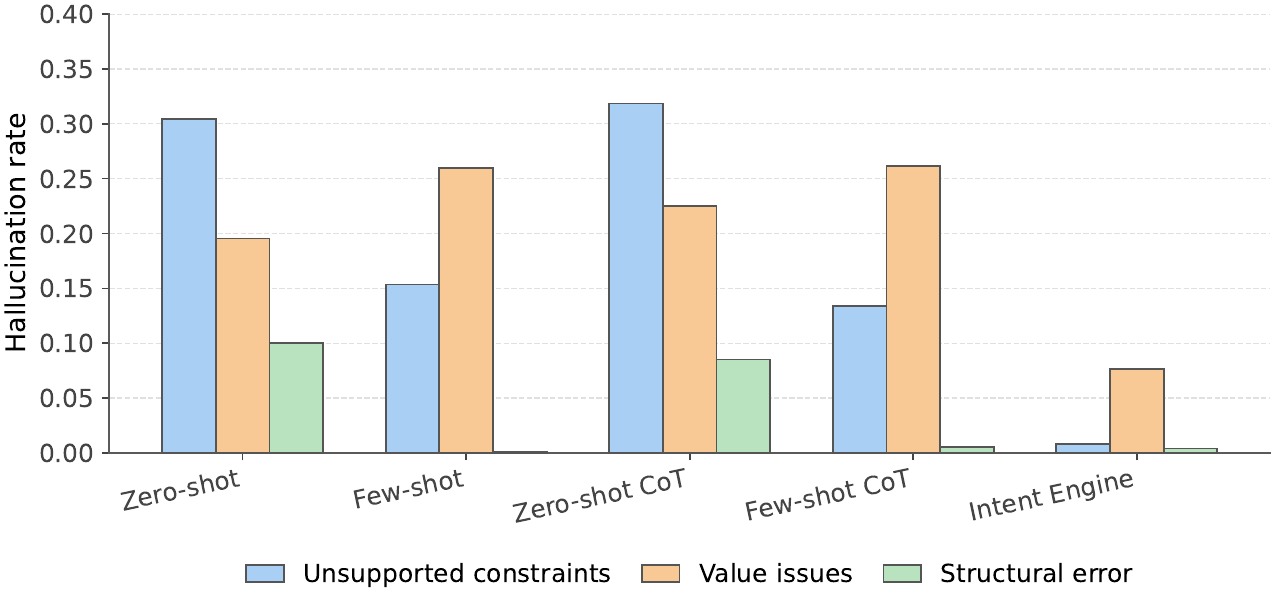}
  \caption{Overall hallucination rates across baselines using GPT-4.1 mini.}
  \label{figure:hallucination_score}
\end{figure*}

Figure~\ref{figure:hallucination_score} and Table~\ref{table:hallucination_breakdown} show that the Intent Engine yields the lowest aggregate hallucination rate across translation methods. Zero-shot prompting produces the highest aggregate score, mainly due to unsupported constraints and incorrect or missing values. Few-shot and CoT prompting reduce some unsupported-constraint and schema errors, but still exhibit value errors.

\begin{table}[!t]
\centering
\caption{Overall hallucination rate by type across baseline translation methods using GPT-4.1 mini.}
\label{table:hallucination_breakdown}

\small
\setlength{\tabcolsep}{4.2pt}
\renewcommand{\arraystretch}{1.15}

\begin{tabular*}{\columnwidth}{@{\extracolsep{\fill}}lcccc@{}}
\hline
\textbf{Method} 
& $\boldsymbol{H_1}$ 
& $\boldsymbol{H_2}$ 
& $\boldsymbol{H_3}$ 
& $\boldsymbol{H}$ \\
\hline
Zero-shot      
& 0.3045 & 0.1955 & 0.1006 & 0.6006 \\

Few-shot       
& 0.1536 & 0.2598 & 0.0014 & 0.4148 \\

Zero-shot CoT  
& 0.3184 & 0.2249 & 0.0852 & 0.6285 \\

Few-shot CoT   
& 0.1341 & 0.2612 & 0.0056 & 0.4008 \\

Intent Engine  
& \textbf{0.0084} & \textbf{0.0768} & 0.0042 & \textbf{0.0894} \\
\hline
\end{tabular*}
\end{table}

\new{Intent Engine reduces aggregate hallucination $H$ by 85.1\% relative to Zero-shot (0.6006$\rightarrow$0.0894). It also reduces unsupported or spurious constraints by 97.2\% ($H_1$: 0.3045$\rightarrow$0.0084), value issues by 60.7\% ($H_2$: 0.1955$\rightarrow$0.0768), and structural errors by 95.8\% ($H_3$: 0.1006$\rightarrow$0.0042). The remaining hallucinations are mostly value-related, showing that grounded value resolution is the hardest part of the SLO construction process.}

\subsection{Retrieval-Grounding Ablation}
\label{rag_ablation_section}

To isolate the effect of retrieval grounding, we remove the RAG module and evaluate the pipeline without snapshot-derived value resolution. In this ablated setting, the system can still extract explicit constraints such as placement requirements and numeric thresholds, but it cannot resolve implicit requirements such as highest or lowest resource/utilization values from the infrastructure state.

\begin{table*}[!t]
\centering
\caption{Retrieval-grounding ablation results for Intent Engine with and without RAG using GPT-4.1 mini model.}
\label{table:rag_ablation}

\small
\setlength{\tabcolsep}{5.5pt}
\renewcommand{\arraystretch}{1.15}

\begin{tabular*}{\textwidth}{@{\extracolsep{\fill}}llccccc@{}}
\hline
\textbf{Variant}
& \textbf{Scope}
& \textbf{L1}
& \textbf{L2}
& \textbf{L3}
& \textbf{L4}
& \textbf{L5} \\
\hline
No RAG
& Explicit + implicit
& 1.000 & 0.926 & 0.775 & 0.604 & 0.500 \\

RAG
& Explicit + implicit
& 1.000 & 0.979 & 0.940 & 0.863 & 0.857 \\
\hline
No RAG
& Implicit only
& -- & 0.000 & 0.000 & 0.000 & 0.000 \\

RAG
& Implicit only
& -- & 0.691 & 0.705 & 0.619 & 0.667 \\
\hline
\end{tabular*}
\end{table*}

\new{Table~\ref{table:rag_ablation} shows that removing retrieval grounding mainly affects higher-complexity intents. When all SLOs are considered, the no-RAG variant still receives partial credit because many high-complexity records combine implicit requirements with explicit placement or numeric-threshold constraints. However, when the evaluation is restricted to implicit snapshot-dependent constraints only, the no-RAG variant collapses to zero F1 across Levels~2--5. With RAG enabled, the Intent Engine recovers nonzero F1 on these implicit constraints, demonstrating that retrieval is essential for resolving grounded highest/lowest values.}

\subsection{Correctness of Generated SLO Specifications}
\label{correctness_generated_slo}

We evaluate correctness at two levels. Contextual accuracy measures whether each generated SLO constraint matches the ground truth after canonicalization of type, metric, operator, value, and unit. Structural accuracy measures whether the complete predicted SLO set matches or overlaps with the ground-truth specification.

\subsubsection{Contextual Accuracy}
\label{contextual_accuracy}

We compute TP, FP, and FN through strict canonical constraint matching and derive Precision, Recall, and F1 using Eqs.~\ref{eq:precision}--\ref{eq:f1}. Table~\ref{table:main_f1_compact} reports the overall F1 Scores across model families and data sources.

\begin{table*}[!t]
\centering
\caption{Overall F1 Score comparison across model families and data sources for baseline methods.}
\label{table:main_f1_compact}

\small
\setlength{\tabcolsep}{4.0pt}
\renewcommand{\arraystretch}{1.18}

\begin{tabular*}{\textwidth}{@{\extracolsep{\fill}}llccccc@{}}
\hline
\textbf{Model} 
& \textbf{Source}
& \textbf{Zero-shot}
& \textbf{Few-shot}
& \textbf{Zero-shot CoT}
& \textbf{Few-shot CoT}
& \textbf{Intent Engine} \\
\hline

\multirow{3}{*}{\shortstack{GPT-4.1\\mini}}
& Real
& 0.296
& 0.711
& 0.555
& 0.716
& \textbf{0.993} \\

& Synthetic
& 0.345
& 0.768
& 0.598
& 0.775
& \textbf{0.920} \\

& \textit{Total}
& 0.331
& 0.751
& 0.585
& 0.758
& \textbf{0.941} \\
\hline

\multirow{3}{*}{\shortstack{Claude\\Sonnet 4.5}}
& Real
& 0.296
& 0.729
& 0.729
& 0.737
& \textbf{0.878} \\

& Synthetic
& 0.348
& 0.777
& 0.822
& 0.795
& \textbf{0.867} \\

& \textit{Total}
& 0.333
& 0.763
& 0.794
& 0.778
& \textbf{0.870} \\
\hline

\multirow{3}{*}{\shortstack{DeepSeek\\V4-Flash}}
& Real
& 0.294
& 0.715
& 0.731
& 0.711
& \textbf{0.869} \\

& Synthetic
& 0.344
& 0.776
& 0.799
& 0.774
& \textbf{0.916} \\

& \textit{Total}
& 0.329
& 0.758
& 0.779
& 0.756
& \textbf{0.903} \\
\hline
\end{tabular*}
\end{table*}

\new{Table~\ref{table:main_f1_compact} shows that the Intent Engine achieves the highest F1 Score for every model family and data source. For GPT-4.1 mini, total F1 increases from 0.758 with the best prompting baseline (Few-shot CoT) to 0.941, an improvement of 24.1\%. The same trend holds for Claude Sonnet 4.5, where F1 increases from 0.794 to 0.870, and for DeepSeek V4-Flash, where F1 increases from 0.779 to 0.903. These correspond to relative improvements of 9.6\% and 15.9\%, respectively, indicating that the gain is not tied to a single backend model.}

\begin{figure*}[t]
  \centering
  \includegraphics[width=1\textwidth]{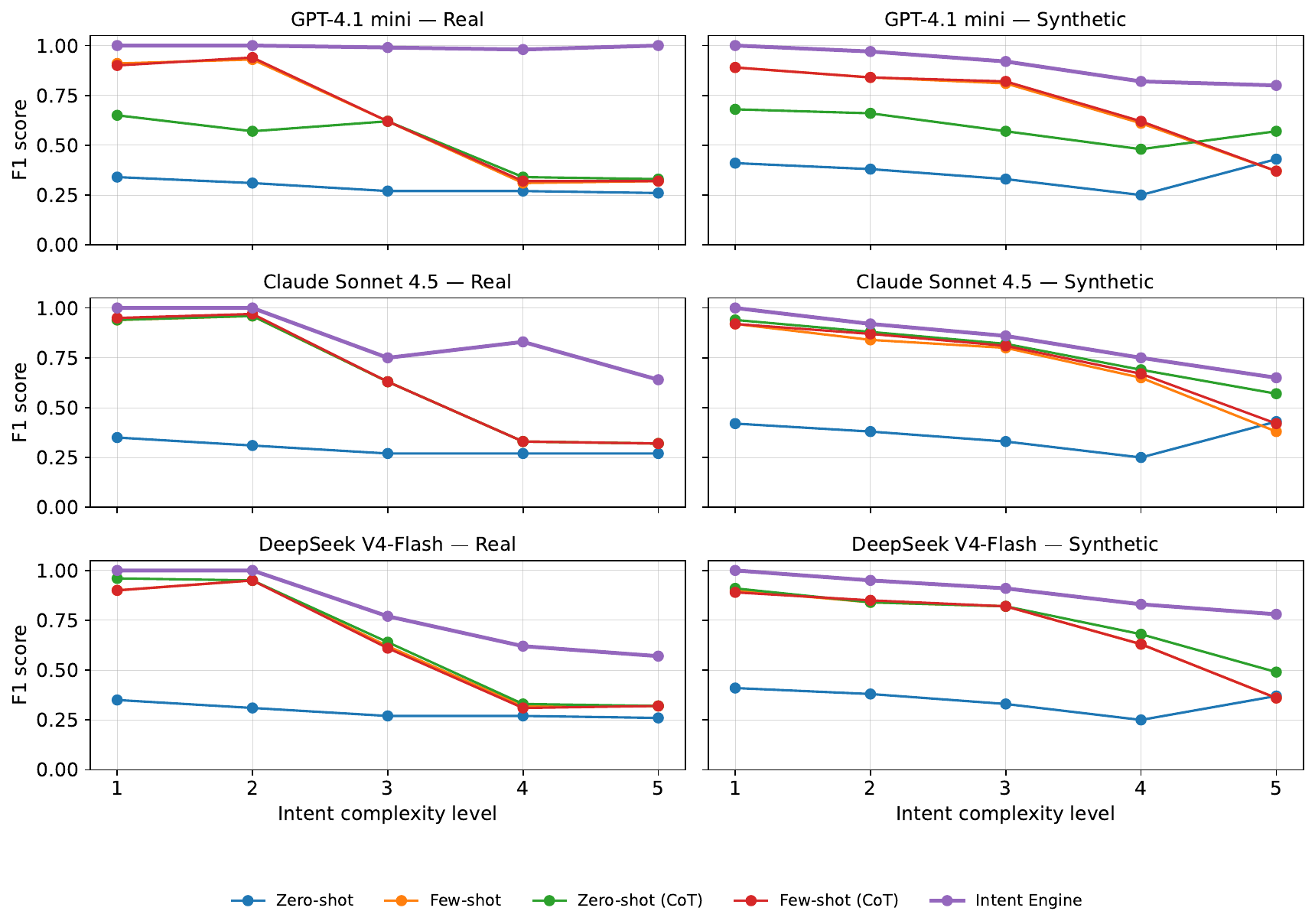}
  \caption{Level-wise F1 Scores across baselines and candidate models for real and synthetic records.}
  \label{figure:f1_scores_across_levels}
\end{figure*}

\new{Figure~\ref{figure:f1_scores_across_levels} shows that prompting baselines degrade as intent complexity increases, especially at higher levels with multiple and implicit constraints. The Intent Engine remains stronger across levels, supporting the benefit of schema-bounded extraction and retrieval grounding for constructing complex orchestration-consumable SLO artifacts.}

\subsubsection{Structural Accuracy}
\label{structural_accuracy}

We evaluate structural fidelity using Exact Match (Eq.~\ref{eq:exactmatch}) and Jaccard similarity (Eq.~\ref{eq:jaccard}). Exact Match requires the entire predicted SLO set to match the ground truth, while Jaccard similarity measures constraint-set overlap.

\begin{figure*}[t]
  \centering
  \includegraphics[width=1\textwidth]{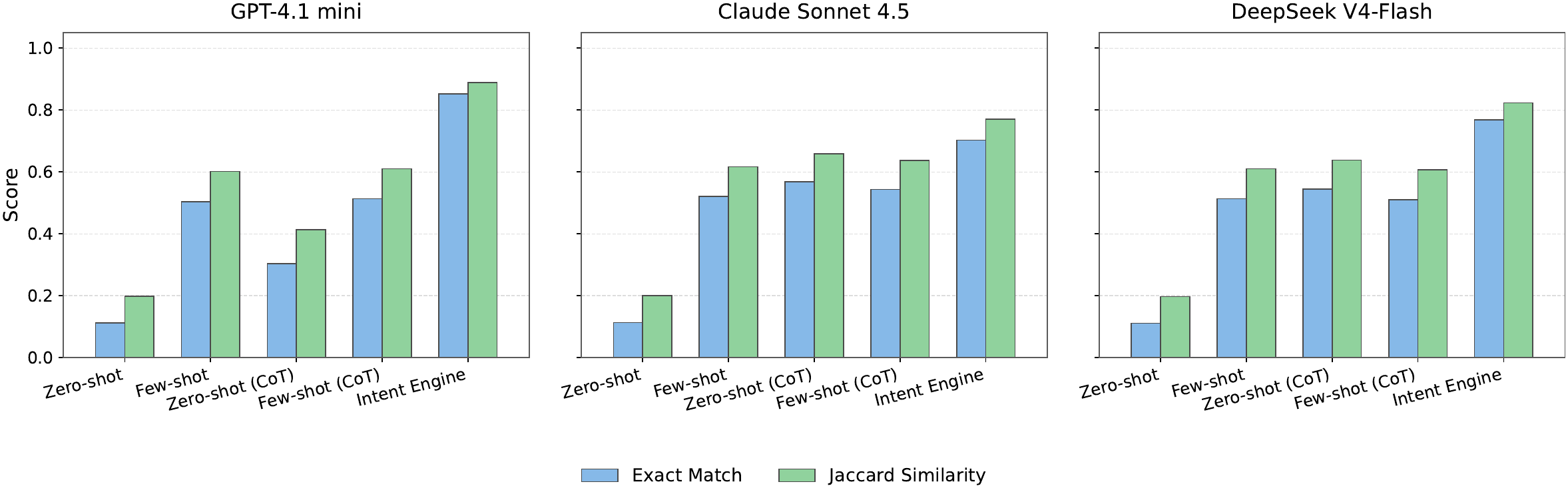}
  \caption{Exact Match and Jaccard similarity across models and translation baselines on the valid evaluation records.}
  \label{figure:exact_match_vs_jaccard}
\end{figure*}

\new{Figure~\ref{figure:exact_match_vs_jaccard} shows that \emph{Intent Engine} also achieves the strongest structural accuracy across models and prompting baselines. This indicates that the improvement is not limited to isolated constraint matches; the generated specifications more often preserve the complete SLO structure, avoid missing or spurious constraints, and remain consistent with the target schema.}

\subsubsection{Rule-based Parser Baseline}
\label{non_llm_structured_parser}

\new{Table~\ref{table:rule_based_f1_compact} compares the rule-based parser with the GPT-4.1 mini prompting baselines and \emph{Intent Engine}. The rule-based parser achieves an overall F1 of 0.666, with similar performance on real and synthetic records. It outperforms zero-shot prompting, but remains below few-shot prompting and \emph{Intent Engine}. This gap highlights the benefit of LLM-based intent interpretation and retrieval-grounded SLO construction for paraphrased and multi-constraint intents.}

\begin{table}[!t]
\centering
\caption{F1 Score comparison with the rule-based parser and prompting baselines using GPT-4.1 mini.}
\label{table:rule_based_f1_compact}

\small
\setlength{\tabcolsep}{10.0pt}
\renewcommand{\arraystretch}{1.15}

\begin{tabular}{lccc}
\hline
\textbf{Method} 
& \textbf{Real}
& \textbf{Synthetic}
& \textbf{Total} \\
\hline
Zero-shot
& 0.296
& 0.345
& 0.331 \\

Zero-shot CoT
& 0.555
& 0.598
& 0.585 \\

Rule-based Parser
& \textbf{0.661}
& \textbf{0.668}
& \textbf{0.666} \\

Few-shot
& 0.711
& 0.768
& 0.751 \\

Few-shot CoT
& 0.716
& 0.775
& 0.758 \\

\emph{Intent Engine}
& \textbf{0.993}
& \textbf{0.920}
& \textbf{0.941} \\
\hline
\end{tabular}
\end{table}

\subsection{Failure Mode Analysis}
\label{failure_modes}

\new{Beyond measuring translation correctness on valid intents, we also evaluate how safely each method handles invalid intents. For such inputs, the desired behavior is not to generate a best-effort SLO, but to reject the request or return no orchestration-consumable SLO specification. We therefore analyze two aspects: (i) representative failure modes in intent-to-SLO translation, and (ii) rejection behavior across invalid categories.}

\new{Table~\ref{table:failure_modes} summarizes the main failure modes observed in the evaluation. The taxonomy distinguishes errors caused by unsupported or underspecified user requests, unconstrained generation, and missing or incorrect contextual grounding.}

\begin{table*}[!t]
\centering
\caption{Failure modes in the intent-to-SLO translation process.}
\label{table:failure_modes}

\small
\setlength{\tabcolsep}{4.0pt}
\renewcommand{\arraystretch}{1.12}

\begin{tabular*}{\textwidth}{@{\extracolsep{\fill}}p{2.6cm}p{5.0cm}p{4.6cm}p{3.5cm}@{}}
\hline
\textbf{Failure mode}
& \textbf{Example intent}
& \textbf{Issue}
& \textbf{System response} \\
\hline

Unsupported
& ``Deploy the service with the lowest carbon footprint.''
& Requests a capability outside the supported SLO schema.
& Reject as unsupported. \\

Incorrect value
& ``Place the service on the node with the lowest memory utilization.''
& Correct metric is detected, but the grounded value is missing or wrong.
& Resolve through retrieval grounding. \\

Structural error
& ``Keep memory utilization reasonable.''
& Output contains malformed fields, units, or operators.
& Reject during schema validation. \\

Ambiguous
& ``Place the service somewhere sensible with good performance.''
& Intent is underspecified and admits multiple interpretations.
& Flag as ambiguous. \\

Conflicting
& ``Run the service on a node with highest memory utilization and lowest memory utilization.''
& Intent contains mutually inconsistent resource-preference constraints.
& Reject as conflicting. \\

Grounding error
& ``Pick the node with the highest available storage.''
& Retrieved or selected contextual value is incorrect or stale.
& Fallback or validation failure. \\

\hline
\end{tabular*}
\end{table*}

\new{Using the rejection rate in Eq.~\ref{eq:rejection_rate}, Figure~\ref{figure:invalid_intent_rejection} reports how often each method correctly avoids producing orchestration-consumable SLOs for invalid intents.}

\new{Figure~\ref{figure:invalid_intent_rejection} shows that the Intent Engine provides the most reliable failure handling across invalid intent categories. Overall, it correctly rejects 98.5\% of invalid records, corresponding to only 3 false accepts out of 195 invalid intents. It achieves higher rejection on ambiguous, malformed, and unsupported intents, and reaches 93.9\% rejection on conflicting intents.

The prompting baselines are less consistent: zero-shot variants reject many ambiguous and unsupported inputs, while few-shot variants often over-generate SLOs for conflicting intents. This suggests that examples improve structured output generation but do not reliably enforce safe rejection.

\begin{center}
  \includegraphics[width=\columnwidth]{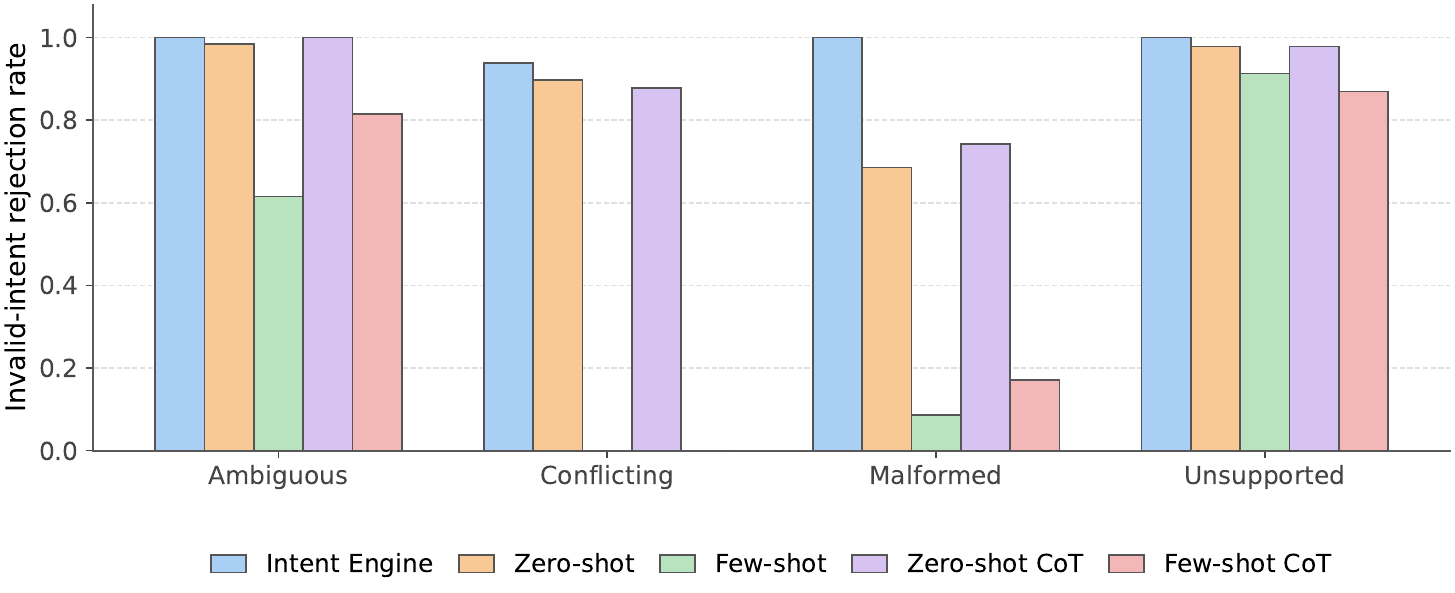}
  \captionof{figure}{Invalid-intent rejection rate across invalid-intent categories using GPT-4.1 mini.}
  \label{figure:invalid_intent_rejection}
\end{center}

These results complement the hallucination analysis: hallucination metrics evaluate errors in generated SLOs, whereas rejection analysis measures whether invalid requests are blocked before validation. Together, they show that \emph{Intent Engine} improves both valid-intent translation and invalid-intent rejection through schema-bounded extraction, retrieval grounding, and deterministic validation.}

\new{
\subsection{System Overhead}
\label{system_overhead}

This section examines the overhead of \emph{Intent Engine} as a translation layer for intent-driven orchestration (IDO) frameworks. The goal is to assess whether the architecture can produce SLO specifications within practical control-plane timescales and whether its context construction remains manageable as the monitored infrastructure grows. We therefore evaluate latency and prompt-context scalability for the intent-to-SLO translation phase.

\subsubsection{Latency}

We measure translation-layer latency from receiving a natural-language intent to producing the final validated SLO specification. Table~\ref{table:intent_engine_latency} reports median, mean, and P95 latency on the total evaluation dataset for the LLM-backed \emph{Intent Engine} configurations.

\begin{table}[!t]
\centering
\caption{Intent Engine runtime latency on the total evaluation dataset. Latency is measured from receiving a natural-language intent to producing the validated SLO specification.}
\label{table:intent_engine_latency}

\small
\setlength{\tabcolsep}{5.0pt}
\renewcommand{\arraystretch}{1.15}

\begin{tabular}{lccc}
\hline
\textbf{Model} 
& \textbf{Median (s)}
& \textbf{Mean (s)}
& \textbf{P95 (s)} \\
\hline
GPT-4.1 mini
& 2.35
& 2.60
& 4.30 \\

Claude Sonnet 4.5
& 4.28
& 4.31
& 5.85 \\

DeepSeek V4-Flash
& 7.78
& 8.76
& 17.43 \\
\hline
\end{tabular}
\end{table}

The results show that \emph{Intent Engine} operates at control-plane translation timescales. GPT-4.1 mini provides the lowest LLM-backed latency, with a median of 2.35~s and P95 of 4.30~s, while DeepSeek V4-Flash has the highest tail latency with a P95 of 17.43~s. For comparison, the rule-based parser completes symbolic parsing and snapshot resolution in sub-millisecond processing time, but with lower F1 than \emph{Intent Engine}. This reflects the expected accuracy--generality trade-off between deterministic parsing and LLM-based intent interpretation. Since intent translation is not performed on the data path of individual service requests, these latencies are suitable for human-triggered or low-frequency orchestration updates where the generated SLOs are consumed by downstream placement or orchestration algorithms.

\subsubsection{Scalability}
\label{scalability}

We evaluate scalability using the LLM-visible context size required for intent-to-SLO translation as the monitored infrastructure grows. The six-node testbed snapshot is scaled to 1000 nodes while preserving the same node-state schema. We compare full snapshot prompting, which inserts the entire continuum state into the prompt, with \emph{Intent Engine}'s retrieval-grounded context construction, which inserts only the top-$k$ metric-relevant evidence chunks. We use $k=8$, matching the evaluated configuration.

Token counts are measured with the GPT-4.1 mini tokenizer. Absolute counts may vary across models, but the architectural trend remains the same: full snapshot prompting grows with infrastructure size, whereas top-$k$ retrieved context is bounded by the retrieval budget.

\begin{table}[t]
\centering
\caption{Prompt-context size with increasing infrastructure scale.}
\label{tab:prompt_context_scalability}
\small
\setlength{\tabcolsep}{6pt}
\renewcommand{\arraystretch}{1.15}
\begin{tabular}{c c c}
\hline
\textbf{Nodes} & \textbf{Full prompt} & \textbf{Retrieved context} \\
\hline
6    & 3,242   & 1,104 \\
25   & 12,694  & 2,012 \\
50   & 25,255  & 2,011 \\
100  & 50,368  & 2,012 \\
250  & 125,600 & 2,012 \\
500  & 250,947 & 2,008 \\
1000 & 501,726 & 2,008 \\
\hline
\end{tabular}
\end{table}

Table~\ref{tab:prompt_context_scalability} shows that full snapshot prompting increases from 3,242 tokens at 6 nodes to 501,726 tokens at 1000 nodes. In contrast, retrieved context stabilizes near 2,000 tokens after 25 nodes because the query retrieves the full $k=8$ evidence budget. For multiple implicit SLOs, retrieved context grows with the number of grounding queries and is bounded by $m \times k$ chunks, where $m$ is the number of implicit grounded constraints. These results show that \emph{Intent Engine} limits LLM-visible context growth while preserving access to infrastructure state for grounding.

}

\new{
\subsection{Downstream Placement Impact}
\label{placement_impact}

To examine whether SLO translation errors affect downstream placement behavior, we perform a placement-impact analysis using the SLO-driven node-matching algorithm from MicroIntent~\cite{islam2025microintent} placement generator. \emph{Intent Engine} remains a platform-agnostic SLO construction layer; the placement generator is used only as a deterministic evaluator to measure how translated SLOs influence a downstream placement decision and accuracy of intended placement intent.

For each valid intent, the placement generator is first applied to the ground-truth SLO label and the corresponding compute-continuum snapshot to check whether a feasible placement exists. The same generator is then applied to each method's predicted SLOs. The generator filters candidate nodes by placement layer, resource constraints, and network constraints, and returns the first node satisfying all SLOs. Placement impact is computed only on the 399 valid records whose ground-truth SLO labels have at least one feasible placement in the snapshot.

\begin{table}[!t]
\centering
\caption{Downstream placement failure using GPT-4.1 mini.}
\label{table:placement_impact}

\small
\setlength{\tabcolsep}{3.5pt}
\renewcommand{\arraystretch}{1.15}

\begin{tabular*}{\columnwidth}{@{\extracolsep{\fill}}lccc@{}}
\hline
\textbf{Method}
& \textbf{No match}
& \textbf{Invalid placement}
& \textbf{Failure} \\
\hline
Zero-shot
& 30.1\%
& 1.5\%
& 31.6\% \\

Few-shot
& 29.3\%
& 1.8\%
& 31.1\% \\

Zero-shot CoT
& 30.1\%
& 1.5\%
& 31.6\% \\

Few-shot CoT
& 29.3\%
& 1.5\%
& 30.8\% \\

Intent Engine
& \textbf{0.8\%}
& \textbf{1.3\%}
& \textbf{2.1\%} \\
\hline
\end{tabular*}
\end{table}

Table~\ref{table:placement_impact} shows that prompting-based translation errors mainly propagate as no-match placement outcomes. The best prompting baseline has a 30.8\% placement failure rate, while the Intent Engine reduces this to 2.1\%. The largest gain comes from reducing no-match cases from 29.3--30.1\% to 0.8\%. Invalid placements remain low because the deterministic placement generator is conservative: incomplete or inconsistent SLOs usually produce no matching node rather than a returned node that violates the ground-truth constraints. These results show that reliable SLO artifact construction reduces the risk of intent-acquisition errors propagating into downstream placement decisions.
}

\new{
\section{Limitations}
\label{limitations}

We designed \emph{Intent Engine} as an intent acquisition and SLO construction layer, not a fully autonomous IDO framework. \emph{Intent Engine} validates and grounds the SLO artifact, while downstream IDO frameworks remain responsible for placement, deployment execution, enforcement, re-grounding, and runtime assurance. For system-critical placement decisions, the generated SLO should be inspected before execution, since stale context, malformed intents, or incorrectly accepted constraints may lead to service disruption.

This work focuses on natural-language-to-SLO construction for compute-continuum service placement. Auto-scaling, fault recovery, migration, cost optimization, and closed-loop assurance are outside the current evaluation. The supported intents are also mainly infrastructure-oriented, covering placement, compute capacity, utilization, storage, and network constraints; business-level goals, privacy, energy, carbon-awareness, application dependencies, and multi-service workflow constraints are not included in the present schema.

The grounding mechanism relies on monitored infrastructure snapshots. Although this resolves implicit SLO values from real system state, the values reflect conditions at snapshot time and may become stale in highly dynamic environments. Thus, the generated SLO is a snapshot-grounded specification rather than an always-current runtime guarantee, while re-grounding after initial placement remains part of the downstream IDO intent assurance phase.

The evaluation uses a six-node edge--cloud testbed and the TeaStore reference application to construct grounded intent-to-SLO records. TeaStore is only a reference microservice application, and the pipeline remains application-agnostic. However, the current setup cannot fully evaluate node-scale behavior or retrieval-augmented context selection in large continuum systems. To the best of our knowledge, no richer public grounded intent-to-SLO dataset exists for this task; once available, \emph{Intent Engine} can be evaluated across larger infrastructures, broader heterogeneity, and more complex deployments.

Finally, \emph{Intent Engine} relies on LLMs for semantic extraction and grounded SLO generation, which can still produce errors despite schema constraints and validation. The invalid-intent rejection and hallucination analyses show that \emph{Intent Engine} reduces unsafe outputs, but does not eliminate all value-level errors. Since no contextual dataset currently exists, training or fine-tuning a domain-specific LLM for accurate SLO construction is not yet feasible.
}

\new{
\section{Conclusion and Future Work}
\label{conclusion_and_future_work}

This paper presented \emph{Intent Engine}, a natural-language-to-SLO construction architecture for compute-continuum service placement. It serves as an intent acquisition layer that transforms unstructured placement intents into validated, orchestration-consumable SLO artifacts for downstream IDO and placement frameworks.

The architecture combines schema-bounded intent extraction, retrieval-grounded value construction, and schema validation to separate natural-language interpretation from downstream orchestration decisions. Evaluation on a real edge--cloud testbed dataset shows that \emph{Intent Engine} improves constraint-level correctness, structural accuracy, hallucination reduction, implicit-value grounding, and invalid-intent rejection compared with prompt-only LLM baselines and a non-LLM rule-based parser. The results show that reliable intent-to-SLO construction requires infrastructure-aware grounding and schema-constrained validation, especially for implicit, multi-constraint, and invalid intents.

Overall, this work shows that natural-language interfaces for compute-continuum orchestration should treat generated SLOs as control-plane artifacts rather than free-form text outputs. By preserving the boundary between intent interpretation and downstream placement execution, \emph{Intent Engine} provides a practical path for integrating natural-language intent acquisition with existing IDO frameworks while leaving placement optimization, deployment actuation, and runtime QoS assurance to the consuming orchestration system.

Future work will investigate intent drift and closed-loop intent assurance after constructed SLOs are consumed by orchestration systems. We plan to extend \emph{Intent Engine} toward a full IDO framework that ingests generated SLO artifacts and verifies runtime intent satisfaction through re-grounding, feasibility checking, conflict detection and resolution, and intent negotiation after initial service placement.
}

\printcredits

\section*{Declaration of Generative AI and AI-assisted technologies in the writing process}

During the preparation of this manuscript, the authors used OpenAI’s ChatGPT for grammar checking and language refinement. All content was reviewed and edited by the authors, who take full responsibility for the accuracy and integrity of the final manuscript.

\section*{Declaration of competing interest}

The authors declare that they have no known competing financial interests or personal relationships that could have influenced the work reported in this paper.

\section*{Acknowledgment}

This work did not receive any grant from funding agencies in the public, commercial, or non-profit sectors.

\section*{Data availability}

The dataset used in this study is publicly available to support reproducibility and further research. The public version of the dataset can be accessed via Zenodo at:

\url{https://doi.org/10.5281/zenodo.20810799}

\appendix

\section{Prompt templates}
\label{appendix:prompt_templates}

\subsection{Intermediate SLO parser prompt template}
\begin{figure*}[!t]
  \centering
  \includegraphics[width=\textwidth]{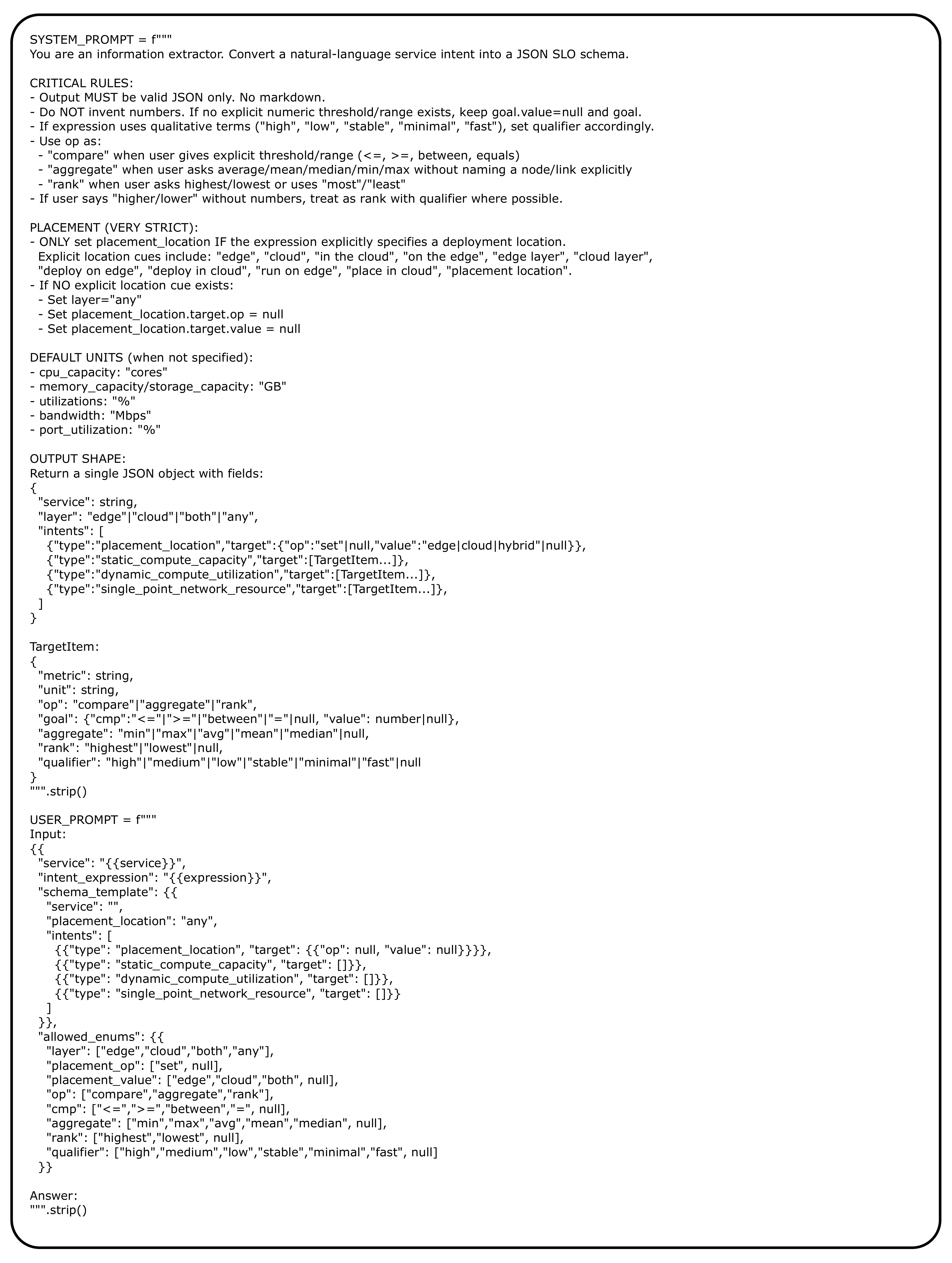}
  \caption{The prompt template used for intent expression decomposition to construct intermediate SLO specification with explicit requirements.}
  \label{figure:parser_prompt_template}
\end{figure*}

See Figure~\ref{figure:parser_prompt_template}.

\subsection{Zero-Shot and Zero-Shot CoT prompt template}
\begin{figure*}[!t]
  \centering
  \includegraphics[width=\textwidth]{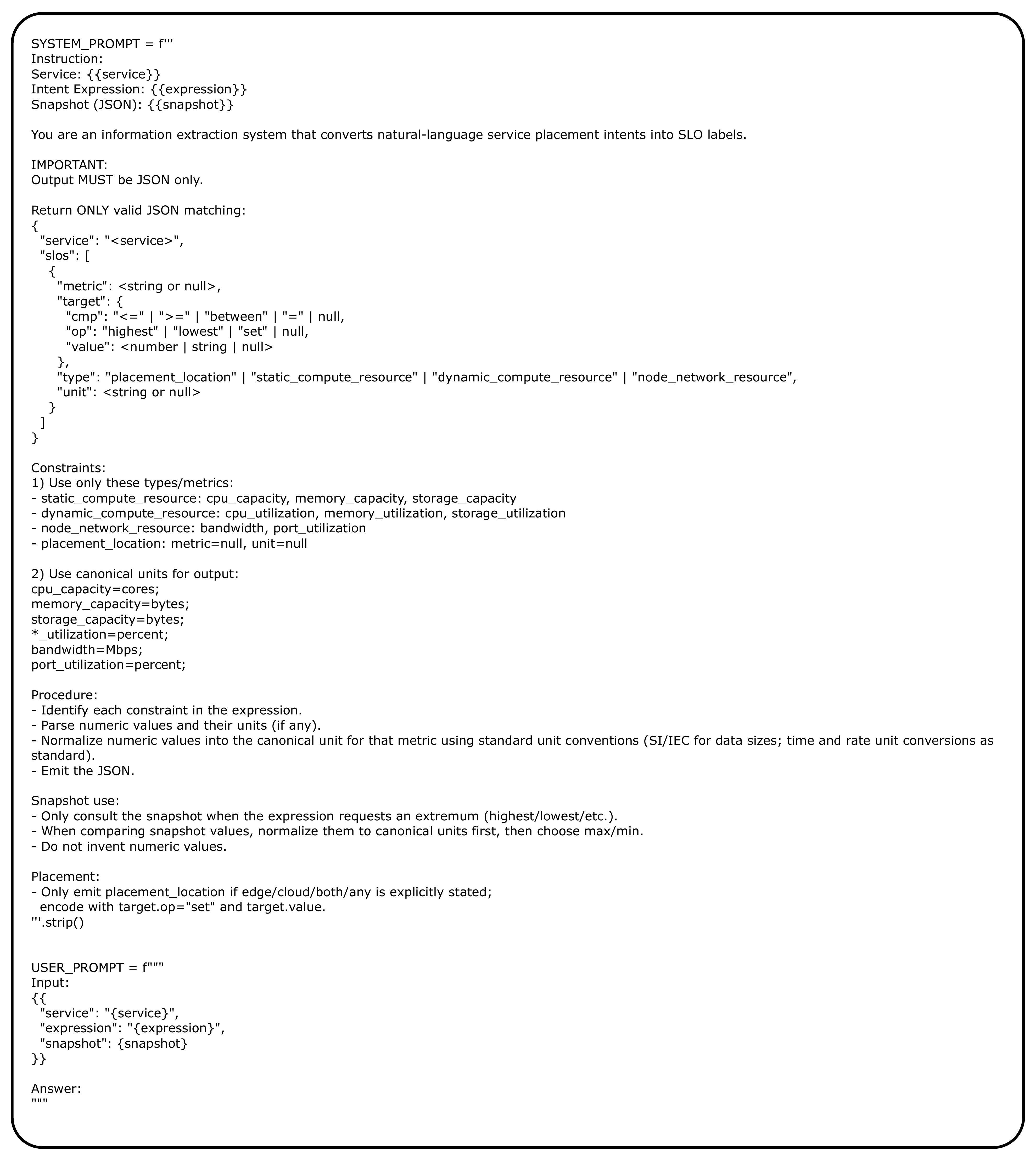}
  \caption{The prompt template used for Zero-Shot learning. For Zero-Shot CoT, we put the phrase ``Let’s think step-by-step'' in the prompt template~\ref{figure:zero_shot_and_cot_prompt_template}.}
  \label{figure:zero_shot_and_cot_prompt_template}
\end{figure*}

See Figure~\ref{figure:zero_shot_and_cot_prompt_template}.

\subsection{Few-Shot and Few-Shot CoT prompt template}
\begin{figure*}[!t]
  \centering
  \includegraphics[width=\textwidth]{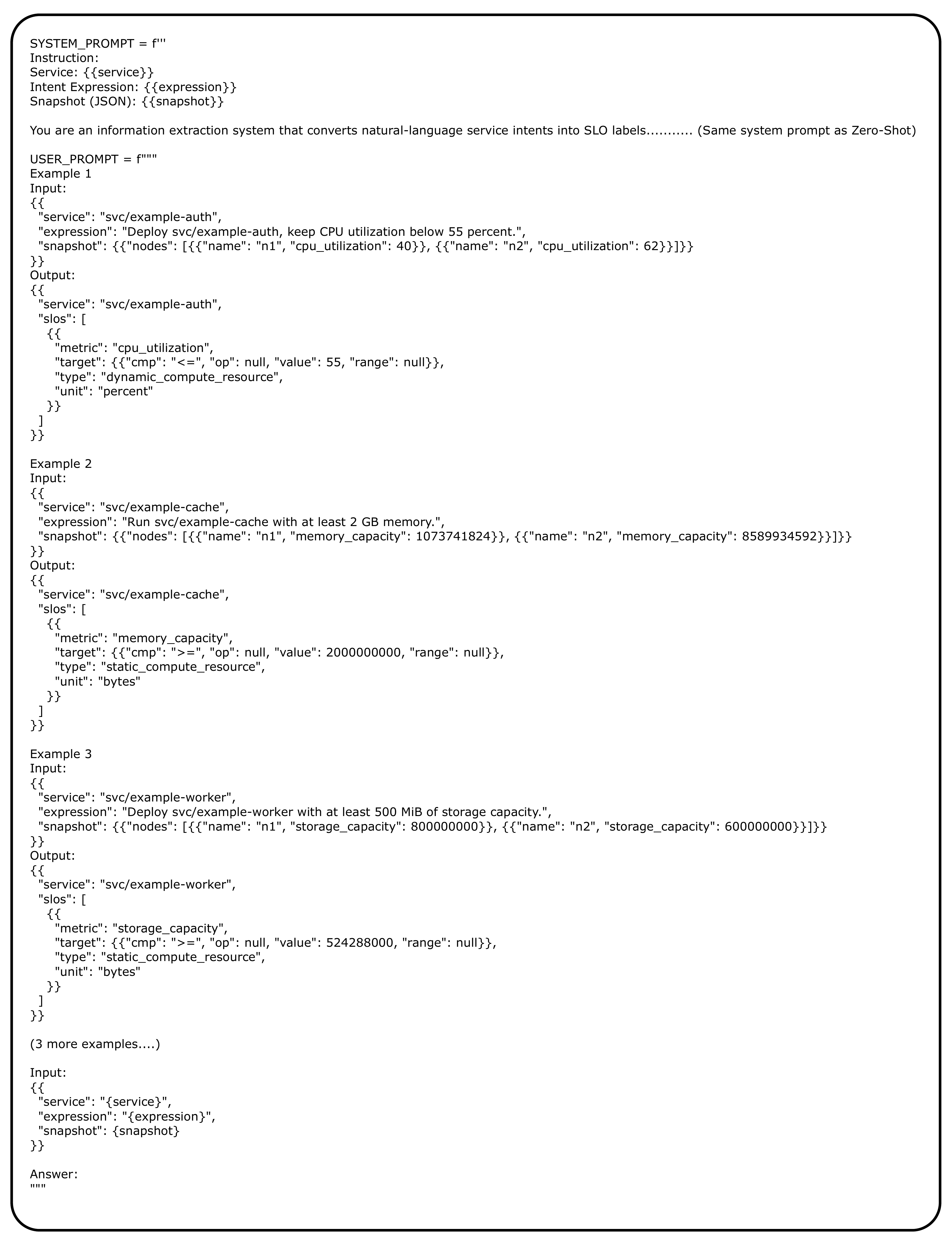}
  \caption{The prompt template used for Few-Shot learning. For Few-Shot CoT, we put the phrase ``Let’s think step-by-step'' is added along with Zero-Shot prompt template~\ref{figure:zero_shot_and_cot_prompt_template} and provided with total 6 examples.}
  \label{figure:fews_shot_and_cot_prompt_template}
\end{figure*}

See Figure~\ref{figure:fews_shot_and_cot_prompt_template}.

\section{Dataset samples}
\label{appendix:dataset_samples}

\subsection{Dataset sample}
\begin{figure*}[!t]
  \centering
  \includegraphics[width=\textwidth]{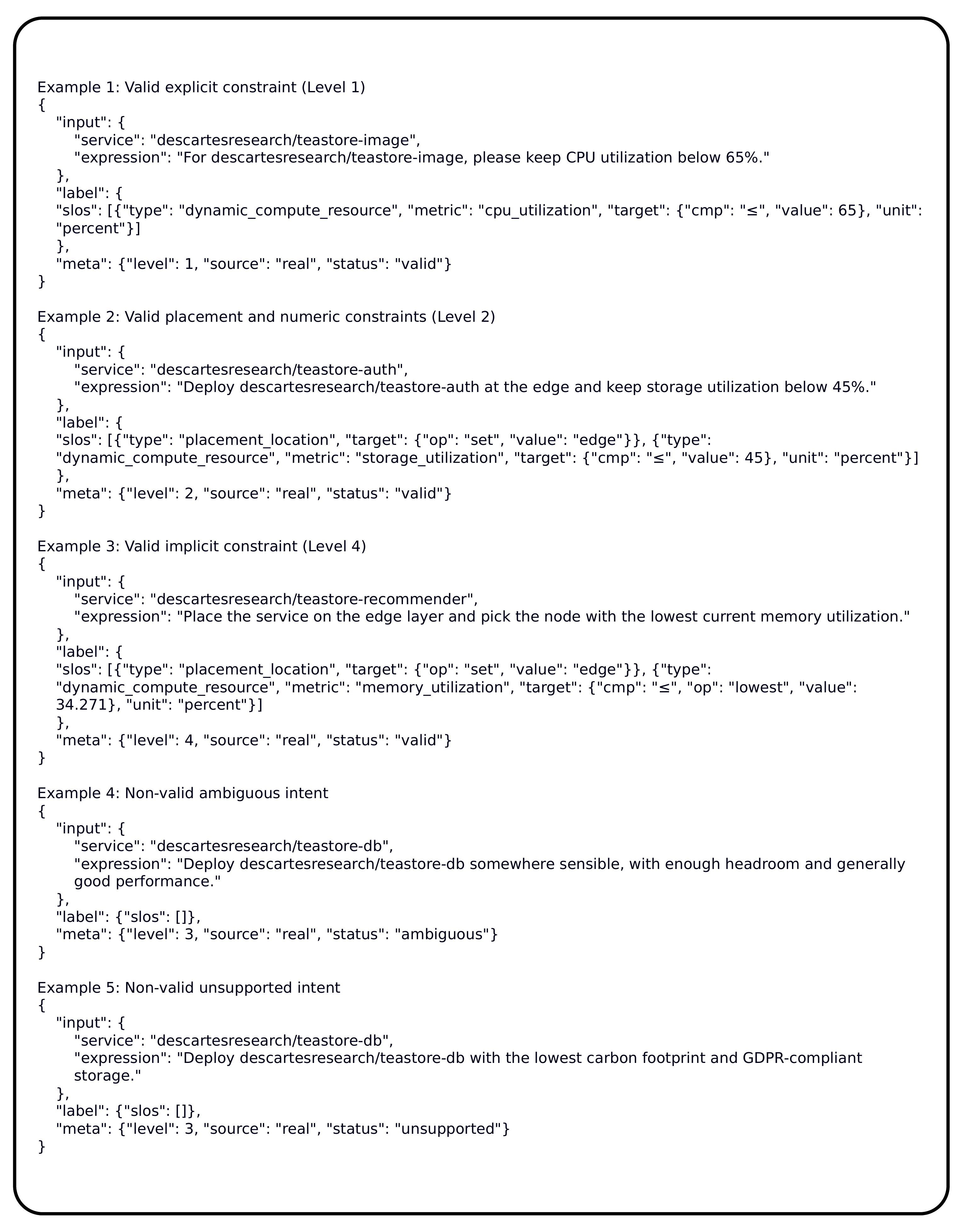}
  \caption{Dataset record examples across valid and invalid natural language intents and ground-truth SLO labels.}
  \label{figure:dataset_sample}
\end{figure*}

See Figure~\ref{figure:dataset_sample}.

\bibliographystyle{model1-num-names}

\bibliography{cas-refs}

\bio{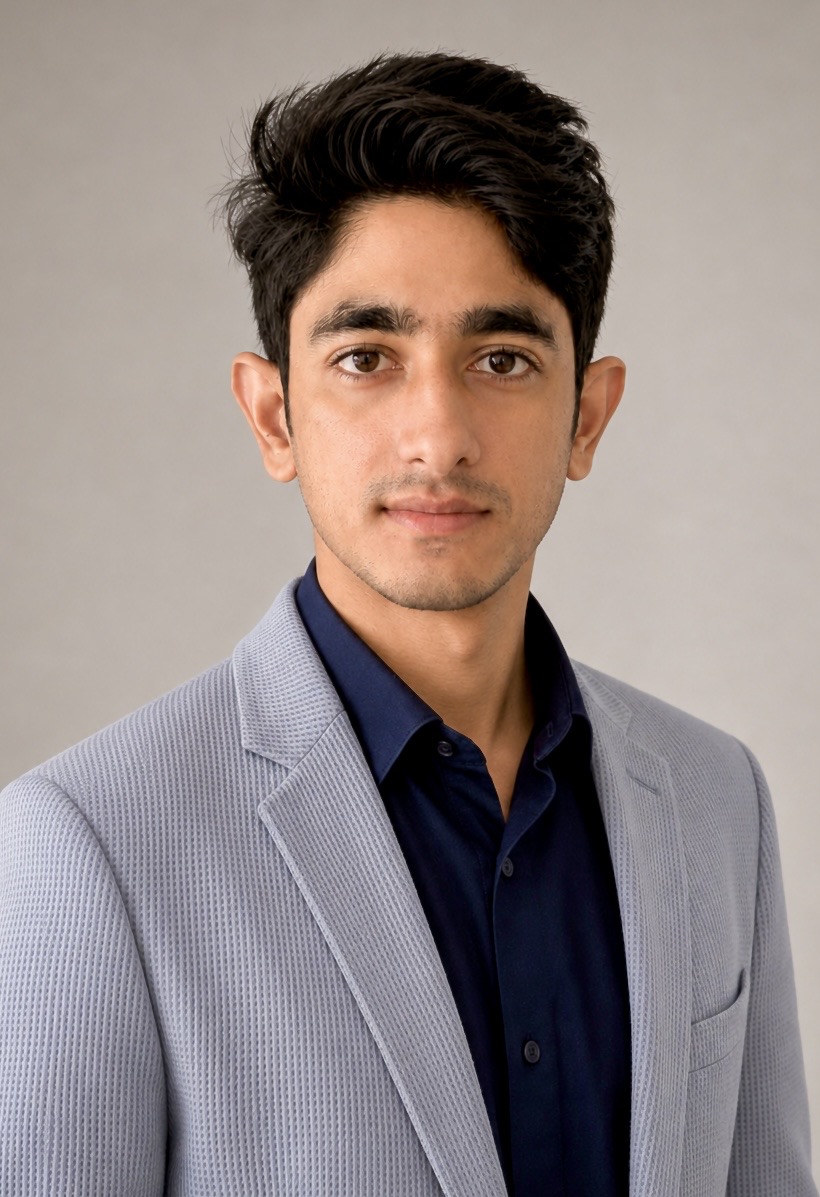}
Koushikur Islam is a Research Assistant at the Smart and Distributed Computing Lab at Western Sydney University, Australia. His research focuses on improving resource management across the edge--cloud continuum through adaptive and autonomous AI/ML-driven solutions. He received his Master of Information and Communications Technology degree from Western Sydney University with High Distinction. He also holds a Bachelor of Science in Computer Science and Engineering from the American International University--Bangladesh, graduating with the Summa Cum Laude (Gold Medal) distinction.
\endbio

\bio{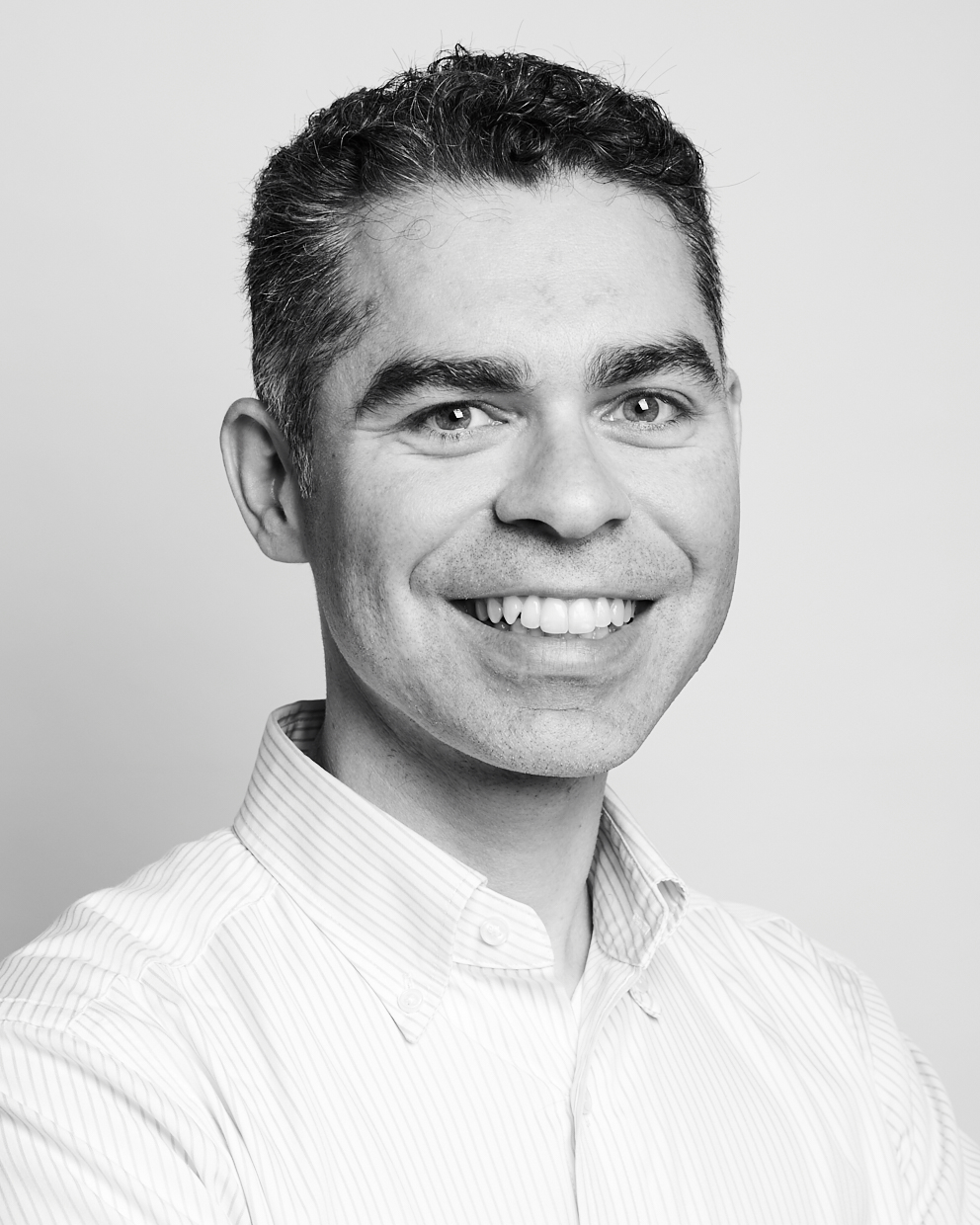}
Dr Rodrigo N. Calheiros is an Associate Professor in the School of Computer, Data and Mathematical Sciences, Western Sydney University, Australia. He conducts applied research in diverse aspects of distributed computing systems, including cloud computing, edge computing, and Internet of Things. He co-authored more than 100 papers, which attracted together 21,000 Google Scholar citations. He is a Fellow of Advance HE, Senior Member of the IEEE and Senior Member of the ACM.
\endbio

\end{document}